\documentclass{article} %
\usepackage{iclr2026_arxiv,times}

\usepackage{amsmath,amsfonts,bm}

\def\eqref#1{equation~\ref{#1}}

\def\1{\bm{1}}

\def\vs{{\bm{s}}}

\def\vu{{\bm{u}}}
\def\vv{{\bm{v}}}

\def\vx{{\bm{x}}}

\def\vz{{\bm{z}}}

\DeclareMathAlphabet{\mathsfit}{\encodingdefault}{\sfdefault}{m}{sl}
\SetMathAlphabet{\mathsfit}{bold}{\encodingdefault}{\sfdefault}{bx}{n}

\usepackage{hyperref}
\usepackage{url}
\usepackage[utf8]{inputenc} %
\usepackage[T1]{fontenc}    %
\usepackage{hyperref}       %
\usepackage{url}            %
\usepackage{booktabs}       %
\usepackage{amsfonts}       %
\usepackage{nicefrac}       %
\usepackage{microtype}      %
\usepackage{xcolor}         %
\usepackage{multirow,diagbox}
\usepackage{graphicx}
\usepackage{subcaption}
\usepackage{amsmath}
\usepackage{comment}
\usepackage[export]{adjustbox} 
\usepackage{amsthm}
\usepackage{amssymb}
\definecolor{SDEblue}{RGB}{28 58 88}
\definecolor{cc1}{rgb}{1.0, 0.44, 0.37}
\definecolor{cc2}{rgb}{0.0, 0.2, 0.6}
\definecolor{cc3}{RGB}{255, 191, 0}
\definecolor{cc4}{RGB}{0, 128, 128}
\hypersetup{
 colorlinks=true,
 urlcolor=magenta,
 citecolor=SDEblue,
}

\renewcommand{\eqref}[1]{Eq.~(\ref{#1})}

\title{Nonparametric Data Attribution for\\Diffusion Models}

\renewcommand\footnotemark{}
\author{Yutian Zhao$^{*1,2}$, Chao Du$^{*\dagger 1}$, Xiaosen Zheng$^{1,3}$, Tianyu Pang$^{1}$, Min Lin$^{1}$
\thanks{$^*$Equal contribution. Work done during Yutian Zhao’s associate membership at Sea AI Lab.}
\thanks{$^{\dagger}$Correspondence to Chao Du.}\\
$^{1}$Sea AI Lab, Singapore \\
$^{2}$Department of Mathematics, National University of Singapore \\
$^{3}$Singapore Management University \\
\footnotesize{\texttt{e0708171@u.nus.edu; \{duchao, tianyupang, linmin\}@sea.com;}}
}

\iclrfinalcopy %
\begin{document}

\maketitle

\begin{abstract}
Data attribution for generative models seeks to quantify the influence of individual training examples on model outputs. Existing methods for diffusion models typically require access to model gradients or retraining, limiting their applicability in proprietary or large-scale settings. We propose a \emph{nonparametric} attribution method that operates entirely on data, measuring influence via patch-level similarity between generated and training images. Our approach is grounded in the analytical form of the optimal score function and naturally extends to multiscale representations, while remaining computationally efficient through convolution-based acceleration. In addition to producing spatially interpretable attributions, our framework uncovers patterns that reflect intrinsic relationships between training data and outputs, independent of any specific model. Experiments demonstrate that our method achieves strong attribution performance, closely matching gradient-based approaches and substantially outperforming existing nonparametric baselines.
\end{abstract}

\section{Introduction}
Recent advances in generative models, particularly diffusion models~\citep{ho2020denoising,song2021scorebased}, have led to significant progress in image synthesis and editing~\citep{rombach2022high,ramesh2022hierarchical,meng2022sdedit}. As these powerful models are trained on increasingly large-scale datasets that often contain private, copyrighted, or low-quality content, concerns around data transparency, accountability, and ethical use are growing~\citep{carlini2023extracting,saveri2023image}. These concerns motivate the study of \emph{data attribution}: identifying the influence of individual training examples on a given generation. Effective data attribution not only supports the responsible deployment of generative models but also enables various downstream applications, including interpreting model behavior~\citep{pmlr-v70-koh17a,NEURIPS2021_c460dc0f,ilyas2022datamodels}, detecting mislabeled or poisoned data~\citep{Jia_2021_CVPR}, guiding data valuation~\citep{nohyun2023data}, and improving dataset quality through informed curation~\citep{Khanna2018InterpretingBB,Jia_2021_CVPR,Liu_2021_ICCV}.
\looseness=-1

Much progress has been made in attributing image generations to training data. Retraining-based methods~\citep{ghorbani2019data,ilyas2022datamodels} assess how generations change when specific training data are removed. While effective, these methods typically require retraining the model tens of thousands of times on different data subsets~\citep{ghorbani2019data}, making them computationally expensive. To improve efficiency, recent works~\citep{zheng2024intriguing,lin2025diffusion,mlodozeniec2025influence} adopt approximations based on additive attribution scores~\citep{pmlr-v202-park23c}, enabling scalable attribution on large datasets. A common assumption in these methods is access to model gradients, i.e., full access to the generative model. However, this assumption is not always practical. For example, in scenarios where users seek copyright protection against infringement by proprietary models~\citep{somepalli2023understanding, zhao2024a}, the model gradients may not be accessible. Furthermore, when attribution is intended to support tasks such as data selection~\citep{gu2025data}, training a generative model solely for attribution may be prohibitively costly or infeasible.

These challenges call for an attribution method that does not require access to the generative model, either to handle proprietary or black-box models or to serve as a fast surrogate without the cost of training or accessing the model.
Existing methods~\citep{zheng2024intriguing,mlodozeniec2025influence}
typically measure the similarity between generated images and training data in feature spaces, but often perform poorly as they disregard the behavior of the generative model.
Effective attribution for diffusion models in this restricted-access setting remains an open challenge.

\begin{figure}[tb]
  \centering
  \includegraphics[
    width=0.95\linewidth]{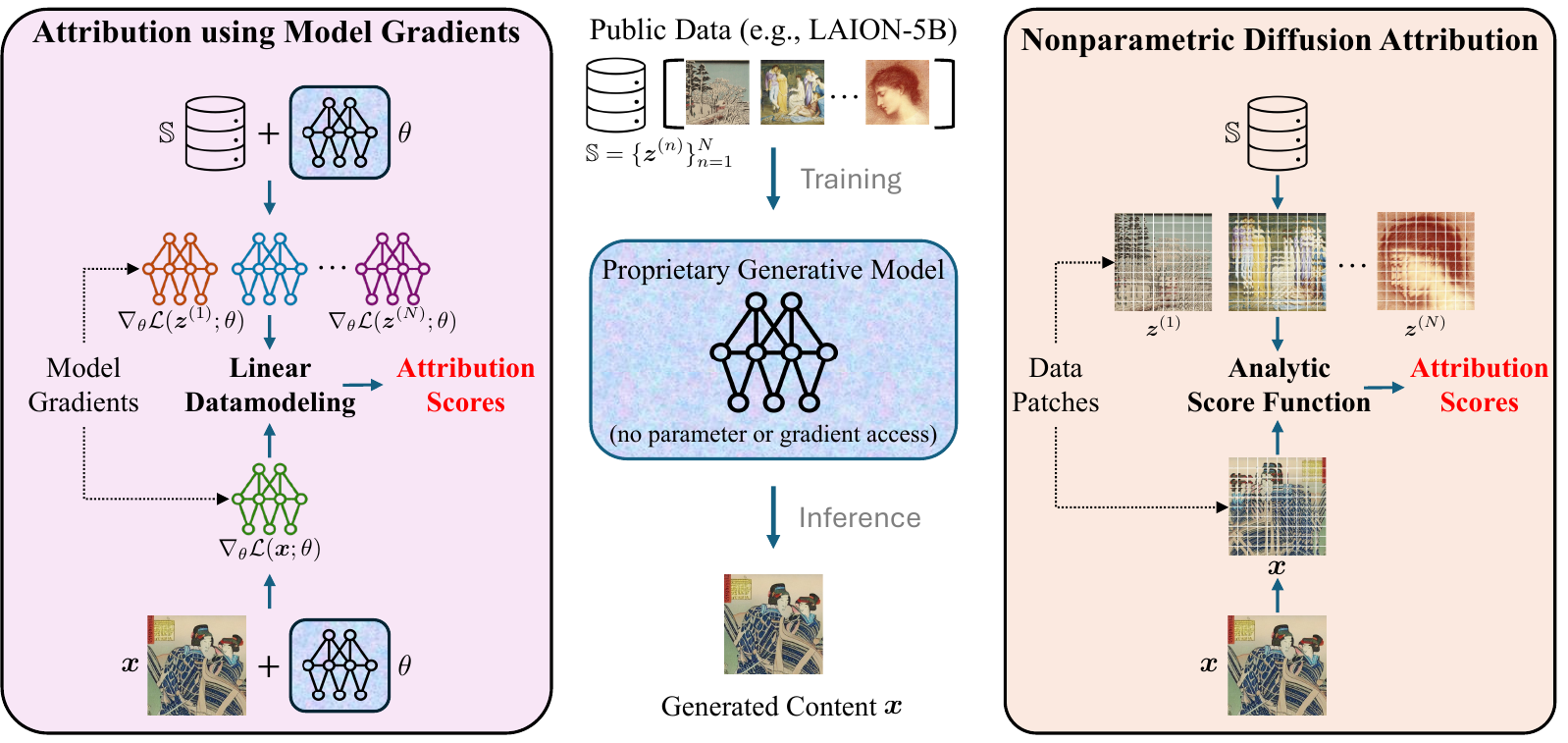}
\caption{Schematic illustration of attribution methods. \textbf{Left:} Model-based attribution relies on gradients and requires parameter access. \textbf{Right:} Our \emph{Nonparametric Diffusion Attribution} (\textbf{NDA}) compares local image patches via an analytic score function, enabling attribution without model access.\looseness=-1}
  \label{fig:score_patch}
\end{figure}

In this paper, we present a \textit{nonparametric} approach to image data attribution in diffusion models, which directly quantifies the influence of training samples on generated outputs through patch-level comparisons. Drawing inspiration from analytical expressions of the score function in diffusion models~\citep{gu2024on,kamb2024analytictheorycreativityconvolutional}, we extract local patches from both generated images and training samples and compute attribution scores based on pairwise distances, as illustrated in Figure~\ref{fig:score_patch}. Unlike prior methods that require retraining or gradient access, our method operates entirely on data, without any reliance on model architecture or training.
Moreover, it provides fine-grained, patch-level attribution, enabling the localization of influential training regions and offering interpretable insights into the generation behaviors.
Due to space constraints, related work is deferred to Appendix~\ref{appendix:related}.\looseness=-1

\section{Preliminaries}
Our approach is grounded in the optimal score function of diffusion models. In this section, we begin with a brief overview of diffusion models and their optimal empirical score functions, followed by a review of data attribution techniques and evaluation protocols.

\subsection{Diffusion models}
Diffusion models~\citep{ho2020denoising,song2021scorebased} are a class of probabilistic generative models that learn to approximate a data distribution $q(\vx)$ by modeling a parametrized Markovian process $p_{\theta}(\vx)$. Specifically, it defines a forward process that transforms a clean data sample $\vx \sim$ $q(\vx)$ into a noisy sequence $\vx_{1: T}=\vx_1, \cdots, \vx_T$ by gradually adding Gaussian noise. The transition probability is given by
$q\left(\vx_t | \vx_{t-1}\right)\triangleq \mathcal{N}\left(\vx_t | \sqrt{1-\beta_{t}} \vx_{t-1}, \beta_{t}\mathbf{I}\right)$, where $\{\beta_{t}\}_{t=1}^{T} $ denotes a predefined variance schedule. A key property of the forward process is that $\vx_{t}$ can be sampled in closed form at any timestep $t$:\looseness=-1
\begin{equation}
q(\vx_{t}|\vx)=\mathcal{N}(\vx_{t}|\sqrt{\bar{\alpha}_{t}}\vx, (1-\bar{\alpha}_{t})\mathbf{I})\textrm{,}
\end{equation}
where $\bar{\alpha}_{t}\triangleq \prod_{s=1}^t \alpha_s$ and $\alpha_{t}\triangleq 1-\beta_{t}$.
As $t$ increases from $0$ to $T$, the noise level increases and $\bar{\alpha}_t$ decays from $1$ to $0$. Consequently, the marginal distribution $q_t(\vx_t)$ approaches a standard Gaussian, i.e., $q_{T}(\vx_{T}) \approx \mathcal{N}(0, \mathbf{I})$.
The reverse process is learned by approximating the reverse conditionals $q(\vx_{t-1}|\vx_{t})$ using a neural network model $p_{\theta}(\vx_{t-1}|\vx_{t}) \triangleq \mathcal{N}(\vx_{t-1}|\boldsymbol{\mu}_{\theta}(\vx_{t}, t), \sigma_{t}^{2}\mathbf{I})$, where the variance $\sigma_{t}^{2}$ are typically chosen as hyperparameters~\citep{bao2022analyticdpm}.
In practice, diffusion models are usually trained to predict the added noise $\boldsymbol{\epsilon}_{\theta}(\vx_{t}, t)$, which relates to the mean via $\boldsymbol{\mu}_{\theta}(\vx_{t}, t) = \frac{1}{\sqrt{\alpha_{t}}}(\vx_{t} - \frac{\beta_{t}}{\sqrt{1 - \bar{\alpha}_{t}}}\boldsymbol{\epsilon}_{\theta}(\vx_{t}, t))$~\citep{ho2020denoising}. 
The model is learned by minimizing a variational bound on the negative log-likelihood:
\begin{equation}
\mathcal{L}_{\text{ELBO}}(\vx; \theta)=\mathbb{E}_{\boldsymbol{\epsilon}, t}\left[\frac{\beta_t^2}{2 \sigma_t^2 \alpha_t\left(1-\bar{\alpha}_t\right)}\left\|\boldsymbol{\epsilon}-\boldsymbol{\epsilon}_\theta\left(\sqrt{\bar{\alpha}_t} \vx+\sqrt{1-\bar{\alpha}_t} \boldsymbol{\epsilon}, t\right)\right\|_2^2\right]\textrm{,}
\end{equation}
where $\boldsymbol{\epsilon} \sim \mathcal{N}(\boldsymbol{\epsilon} | \mathbf{0}, \mathbf{I})$. Let $\mathbb{S} \triangleq \{\vz^{(n)}|\vz^{(n)}\sim q\}_{n=1}^N$ be the training dataset. The empirical training objective is then $\mathcal{L}_{\text{ELBO}}(\mathbb{S}; \theta)=\frac{1}{N} \sum_{n=1}^{N} \mathcal{L}_{\text{ELBO}}\left(\vz^{(n)}; \theta\right)$.
To enhance sample quality, a simplified training objective is often used~\citep{ho2020denoising}:
\begin{equation}
\mathcal{L}_{\text{Simple}}(\vx; \theta)=\mathbb{E}_{\boldsymbol{\epsilon}, t}\left[\left\|\boldsymbol{\epsilon}_\theta\left(\vx_t, t\right)-\boldsymbol{\epsilon}\right\|^2\right]\textrm{,}
\label{eq:simple_objective}
\end{equation}
with the empirical objective on $\mathbb{S}$ given by $\mathcal{L}_{\text {Simple}}(\mathbb{S} ; \theta)=\frac{1}{N} \sum_{n=1}^{N} \mathcal{L}_{\text{Simple}}(\vz^{(n)}; \theta)$.

\subsection{Optimal score functions}
A key property~\citep{song2021scorebased,kingma2023understanding} of diffusion models is that the optimal noise prediction function $\boldsymbol{\epsilon}^*(\vx_t, t)$ is closely related to the \emph{score function} $\vs(\vx_t, t) \triangleq \nabla_{\vx_t}\log q_t(\vx_t)$ via:
\begin{equation}
\vs(\vx_t, t) = -\dfrac{\boldsymbol{\epsilon}^*(\vx_t, t)}{\sqrt{1-\bar{\alpha}_{t}}}.
\end{equation}
Intuitively, for a finite dataset $\mathbb{S} = \{\vz^{(n)}\}_{n=1}^N$, the marginal distribution $q_t\left(\vx_t\right)$ becomes a Gaussian mixture centered at the scaled data points $\sqrt{\bar{\alpha}_t} \vz^{(n)}$: 
$
q_{t}(\vx_{t})=\frac{1}{N} \sum_{n=1}^{N} \mathcal{N}\left(\vx_{t} | \sqrt{\bar{\alpha}_t} \vz^{(n)},\left(1-\bar{\alpha}_t\right) \mathbf{I}\right)
$.
The score function $\vs(\vx_{t}, t)$ therefore admits an analytical form:
\begin{equation}\label{eq:opt_score}
\vs(\vx_{t}, t) = \nabla_{\vx_t}\log q_t(\vx_t) = \dfrac{1}{1-\bar{\alpha}_{t}}\sum_{n=1}^N(\sqrt{\bar{\alpha}_{t}}\vz^{(n)} - \vx_{t})W_{t}(\vz^{(n)}|\vx_{t})\textrm{,}
\end{equation}
where $W_{t}(\vz^{(n)}|\vx_{t})$ is a weighting term defined as:
\begin{equation}\label{eq:opt_score_wt}
W_t(\vz^{(n)} | \vx_{t})=\frac{\mathcal{N}\left(\vx_{t} |\sqrt{\bar{\alpha}_t} \vz^{(n)},\left(1-\bar{\alpha}_t\right) \mathbf{I}\right)}{\sum_{n'=1}^{N} \mathcal{N}\left(\vx_{t} | \sqrt{\bar{\alpha}_t} \vz^{(n')},\left(1-\bar{\alpha}_t\right) \mathbf{I}\right)}.
\end{equation}
This score function $\vs(\vx_{t}, t)$ can be interpreted as a conditional average over the added noise, where the residual term $\vx_{t} - \sqrt{\bar{\alpha}_{t}}\vz^{(n)}$ is averaged across training examples, weighted by the posterior probability $W_{t}(\vz^{(n)}|\vx_{t})$ that $\vx_{t}$ was transformed from $\vz^{(n)}$ at $t=0$ under the forward process.

\subsection{Data attribution and evaluation metrics}
Given a training dataset $\mathbb{S}$ and a generated sample $\vx$, the goal of data attribution is to quantify the influence of each training example in $\mathbb{S}$ on the generation of $\vx$. Formally, this involves assigning an attribution score $\tau(\vx,\vz^{(n)};\mathbb{S})$ to each training example, reflecting its relative importance in generating $\vx$.
\looseness=-1

We follow \citet{zheng2024intriguing} and adopt the linear datamodeling score~(LDS; \citealp{pmlr-v202-park23c}) as our evaluation metric, which quantifies how well an attribution method aligns with the ground-truth influence of training data on model outputs. Specifically, for a training dataset $\mathbb{S}=\{\vz^{(n)}\}^N_{n=1}$ of size $N$, LDS evaluates an attribution method $\tau$ by first sampling multiple random subsets $\{\mathbb{S}_m \subset \mathbb{S}\}_{m=1}^M$.
Let $\theta^*(\mathbb{S}_m)$ denote the generative model trained on subset $\mathbb{S}_m$, and let $\boldsymbol{\mathcal{F}}(\vx;\theta^{*}(\mathbb{S}_m))$ denote the model output on a test input $\vx$.\footnote{For diffusion models, we set $\boldsymbol{\mathcal{F}}=\mathcal{L}_{\text{Simple}}$, representing the model output used in LDS evaluation.}
Then, for each subset $\mathbb{S}_m$, the attribution scores $\tau(\vx,\vz^{(n)};\mathbb{S})$ assigned to training examples in $\mathbb{S}_m$ are summed to form an attribution-based prediction $g_{\tau}(\vx,\mathbb{S}_m;\mathbb{S})$:
\begin{equation}
    g_{\tau}(\vx,\mathbb{S}_m;\mathbb{S}) \triangleq \sum_{\vz^{(n)}\in \mathbb{S}_m}\tau(\vx,\vz^{(n)};\mathbb{S})\textrm{.}
\end{equation}
Finally, the LDS score for attribution method $\tau$ on test input $\vx$ is calculated as the Spearman rank correlation $\rho(\cdot,\cdot)$ between the ground-truth outputs from the $M$ retrained models (each trained on a different subset $\mathbb{S}_m$) and the corresponding attribution-based predictions:
\begin{equation}
{\text{LDS}}(\tau, \vx) \triangleq \rho\Big(\{\boldsymbol{\mathcal{F}}(\vx;\theta^{*}(\mathbb{S}_m)): m\in [M]\}, \{g_{\tau}(\vx,\mathbb{S}_m;\mathbb{S}): m\in [M]\}\Big)\textrm{.}
\label{eq:lds}
\end{equation}

\section{Methodology}
Our approach is motivated by the weighting term in the optimal score function of diffusion models, which naturally encodes the relative importance of each training example during generation. Leveraging this insight, we build upon the recent theoretical framework of \citet{kamb2024analytictheorycreativityconvolutional}, which extends the analysis of optimal score functions to capture inductive biases that promote generalization in diffusion models. By bridging this framework with the problem of data attribution, we develop a nonparametric data attribution method that does not require access to model gradients or retraining. %

\subsection{Equivariant and local score machines}
While \eqref{eq:opt_score} defines the optimal score function, it relies solely on distance-based weighting in image space and does not generalize beyond the training set. \citet{kamb2024analytictheorycreativityconvolutional} derive an analytical form under inductive biases of \emph{locality} and \emph{equivariance}. This formulation preserves the structure of the empirical score but yields meaningful similarity by incorporating spatial structure and symmetry.
\looseness=-1

Let $\vx_{t} \in \mathbb{R}^{C \times L \times L}$ denote a noisy image at diffusion time $t$, with $C$ channels and spatial resolution~$L$.
For a pixel location $\ell\in[L]\times[L]$, let $\vx_{t, \ell}\in\mathbb{R}^C$ denote its pixel value. Define $\Omega_{\ell}$ as the $P \times P$ neighborhood centered at $\ell$, and $\vx_{t,\Omega_\ell} \in \mathbb{R}^{C \times P \times P}$ as the corresponding local patch.
Let $\mathbb{P}_\Omega(\mathbb{S})$ be the set of all such patches extracted from the training dataset $\mathbb{S}$. Each patch $\vu \in \mathbb{P}_\Omega(\mathbb{S})$ is thus a local crop of some training image $\vz \in \mathbb{S}$, and we denote its center pixel as $\vu_0$.
Under the locality and equivariance assumptions, \citet{kamb2024analytictheorycreativityconvolutional} show that the optimal MMSE estimator of the score function at pixel location $\ell$, denoted by $\vs(\vx_t, t, \ell)\in\mathbb{R}^C$, takes the form:
\begin{equation}
\vs(\vx_t, t, \ell) = \sum_{\vu \in \mathbb{P}_{\Omega}(\mathbb{S})} 
\dfrac{\sqrt{\bar{\alpha}_{t}}\vu_0 -\vx_{t, \ell}}{1-\bar{\alpha}_{t}}
W_{t}(\vu| \vx_{t,\Omega_\ell})\textrm{,}
\label{eq:local_score}
\end{equation}
where the weighting term $W_{t}(\vu | \vx_{t,\Omega_\ell})$ is defined as:
\begin{equation}
W_{t}(\vu | \vx_{t,\Omega_\ell}) = \frac{\mathcal{N}\left(\vx_{t, \Omega_\ell} | \sqrt{\bar{\alpha}_t} \vu,\left(1-\bar{\alpha}_t\right) \mathbf{I}\right)}{\sum_{\vv \in \mathbb{P}_{\Omega}(\mathbb{S})} \mathcal{N}\left(\vx_{t, \Omega_\ell} | \sqrt{\bar{\alpha}_t} \vv,\left(1-\bar{\alpha}_t\right) \mathbf{I}\right)}\textrm{.}
\label{eq:local_weight}
\end{equation}
This formulation generalizes \eqref{eq:opt_score} by measuring similarity at the patch level rather than over full images, allowing it to exploit fine-grained local structure.

\subsection{Patch-based data attribution scores}
The weighting term in \eqref{eq:local_weight} can be expressed as a softmax over quadratic distances $\frac{\|\vx_{t,\Omega_\ell} - \sqrt{\bar{\alpha}_{t}} \vu\|^{2}}{2(1-\bar{\alpha}_{t})}$, which naturally reflects the contribution of each training patch $\vu$ to the generation of pixel $\vx_{t, \ell}$. We reinterpret this term as a \emph{patch-wise influence} score that quantifies the influence of a training patch $\vu$ on the local region of a generated image. By aggregating these local scores across spatial locations, we obtain a nonparametric, spatially interpretable attribution measure.

\textbf{Patch-wise influence.} For a noisy patch $\vx_{t,\Omega_\ell}$ centered at $\ell$ in a generated image $\vx$ at timestep $t$, we adopt the local weighting from \eqref{eq:local_weight} and define the patch-wise influence score as:
\begin{equation}
\!\!\tau(\vx_{t,\Omega_\ell}, \vu;\mathbb{P}_{\Omega}(\mathbb{S})) \triangleq\exp\!\left(-\frac{\left\| \vx_{t, \Omega_\ell} \!-\! \sqrt{\bar{\alpha}_t} \vu \right\|^2 }{2(1 - \bar{\alpha}_t)}\right)\!\cdot\!\left( \sum_{\vv \in \mathbb{P}_{\Omega}(\mathbb{S})}\!\exp\!\left(-\frac{\left\| \vx_{t, \Omega_\ell} \!-\! \sqrt{\bar{\alpha}_t} \vv \right\|^2 }{2(1 - \bar{\alpha}_t)}\right)\!\right)^{\!-1}\!\textrm{,}
\label{eq:patch_softmax}
\end{equation}
which is a normalized similarity score over all patches in $\mathbb{P}_{\Omega}(\mathbb{S})$ extracted from the training set.

\textbf{Image-level attribution.}
Given a generated image $\vx$, we apply the forward diffusion process to obtain noisy samples $\vx_t$. For each pixel location $\ell$, we extract the $P \times P$ patch $\vx_{t,\Omega_\ell}$ from $\vx_t$, using zero-padding near boundaries to handle incomplete patches. The patch-wise influence scores in \eqref{eq:patch_softmax} is then computed for all training patches $\vu\in\mathbb{P}_{\Omega}(\mathbb{S})$.

To aggregate into an image-level attribution score $\tau(\vx, \vz^{(n)};\mathbb{S})$, we proceed as follows:
(1)~For each patch $\vx_{t,\Omega_\ell}$ of $\vx_t$, we select the $k$ most influential patches from each training image $\vz^{(n)}$, denoted by $\mathbb{P}^{k}_\Omega(\vx_{t,\Omega_\ell},\vz^{(n)})$.\footnote{Formally, $\mathbb{P}^{k}_\Omega(\vx_{t,\Omega_\ell},\vz^{(n)}) \subset \mathbb{P}_\Omega(\{\vz^{(n)}\})$ with $|\mathbb{P}^{k}_\Omega(\vx_{t,\Omega_\ell},\vz^{(n)})|=k$, and for all $\vu'\in\mathbb{P}^{k}_\Omega(\vx_{t,\Omega_\ell},\vz^{(n)})$ and $\vu''\in \mathbb{P}_\Omega(\{\vz^{(n)}\})\setminus \mathbb{P}^{k}_\Omega(\vx_{t,\Omega_\ell},\vz^{(n)})$, we have $\tau(\vx_{t,\Omega_\ell},\vu';\mathbb{P}_{\Omega}(\mathbb{S}))\geq\tau(\vx_{t,\Omega_\ell},\vu'';\mathbb{P}_{\Omega}(\mathbb{S}))$.}
(2)~We sum the influence scores of these top-$k$ patches to estimate the contribution of the training image $\vz^{(n)}$ to generating the local region around $\ell$.
(3)~Finally, we aggregate across all spatial locations $\ell$ and average over a set of timesteps $\mathcal{T}$:
\begin{equation}
    \tau(\vx,\vz^{(n)}; \mathbb{S})
\triangleq \frac{1}{|\mathcal{T}|}\sum_{t\in\mathcal{T}}\sum_{\ell}
\sum_{\vu \in \mathbb{P}^k_\Omega(\vx_{t,\Omega_\ell},\vz^{(n)})}
\tau(\vx_{t,\Omega_\ell}, \vu; \mathbb{P}_{\Omega}(\mathbb{S}))\textrm{.}
\end{equation}
This attribution score is both \emph{spatially interpretable}, as it aggregates patch-wise influence with local meaning, and \emph{nonparametric}, as it operates entirely on training data without relying on the model.
\looseness=-1

\subsection{Attribution with multiscale patch-wise influence}

The attribution method described above relies on quadratic Euclidean distances between fixed-size patches in the original image space. However, images may have varying resolutions, and a single patch size may fail to capture different levels of information: from fine-grained textures to higher-level structures~\citep{adelson1984pyramid,lin2017feature}. Moreover, diffusion models are known to generate different levels of detail across timesteps: early (high-noise) stages capture coarse structures, while later (low-noise) stages refine local details~\citep{jing2022subspace,park2023understanding}.

To account for these effects, we introduce a \emph{multiscale} extension that computes patch-wise influence across multiple resolutions. Specifically, we downsample both generated and training patches and evaluate distances in the lower-resolution space. Let $\text{D}(\cdot)$ denote a downsampling operator, and define $\widehat{\vx}_{t,\Omega_\ell} \triangleq \text{D}(\vx_{t,\Omega_\ell})$, $\widehat{\vu} \triangleq \text{D}(\vu)$, and $\widehat{\vv} \triangleq \text{D}(\vv)$. The low-resolution patch-wise influence score is then:
\begin{equation}
\!\!\widehat{\tau}(\vx_{t,\Omega_\ell}, \vu;\mathbb{P}_{\Omega}(\mathbb{S})) \triangleq\exp\!\left(-\frac{\left\| \widehat{\vx}_{t, \Omega_\ell} \!-\! \sqrt{\bar{\alpha}_t} \widehat{\vu} \right\|^2 }{2(1 - \bar{\alpha}_t)}\right)\!\cdot\!\left( \sum_{\vv \in \mathbb{P}_{\Omega}(\mathbb{S})}\!\exp\!\left(-\frac{\left\| \widehat{\vx}_{t, \Omega_\ell} \!-\! \sqrt{\bar{\alpha}_t} \widehat{\vv} \right\|^2 }{2(1 - \bar{\alpha}_t)}\right)\!\right)^{\!-1}\!\textrm{.}
\label{eq:patch_softmax_down}
\end{equation}
We then combine the original and low-resolution influence measures into a multiscale score:
\begin{equation}
\label{eq:patch_softmax_ms}
\tau^{\text{ms}}(\vx_{t,\Omega_\ell}, \vu;\mathbb{P}_{\Omega}(\mathbb{S}))
\triangleq
\gamma_t\tau(\vx_{t,\Omega_\ell}, \vu;\mathbb{P}_{\Omega}(\mathbb{S}))
+(1-\gamma_t)\widehat{\tau}(\vx_{t,\Omega_\ell}, \vu;\mathbb{P}_{\Omega}(\mathbb{S}))\textrm{,}
\end{equation}
where $\gamma_t \in [0,1]$ is a timestep-dependent weighting factor that balances the contribution of the original and low-resolution influence.
Finally, we extend this to multiscale image-level attribution by aggregating over timesteps $\mathcal{T}$ and spatial locations $\ell$:
\begin{equation}
\label{eq:image_ms}
\tau^{\text{ms}}(\vx, \vz^{(n)}; \mathbb{S})
\triangleq \frac{1}{|\mathcal{T}|}\sum_{t\in\mathcal{T}}\sum_{\ell}
\sum_{\vu \in \mathbb{P}^k_{\Omega}(\vx_{t,\Omega_\ell}, \vz^{(n)})}
\tau^{\text{ms}}(\vx_{t,\Omega_\ell}, \vu; \mathbb{P}_{\Omega}(\mathbb{S})) \textrm{.}
\end{equation}

\subsection{Convolution-based acceleration}

Directly evaluating Eq.~(\ref{eq:patch_softmax} \& \ref{eq:patch_softmax_down}) on large-scale datasets is computationally challenging. In a naive implementation, each test patch $\vx_{t,\Omega_\ell}$ is broadcast against all unfolded training patches $\vu\in\mathbb{P}_{\Omega}(\mathbb{S})$, leading to peak memory consumption of $\mathcal{O}(N L^2 C P^2)$, which is $P^2$ times larger than the dataset itself.
\looseness=-1

To address this, we propose a memory-efficient implementation that avoids explicit patch unfolding by leveraging convolutional operators. Specifically, to compute $\|\vx_{t,\Omega_\ell} - \sqrt{\bar{\alpha}_{t}} \vu\|^{2}$ for all $L \times L$ patches $\vu$ from a training image $\vz^{(n)}$, we treat the patch $\vx_{t,\Omega_\ell} \in \mathbb{R}^{C \times P \times P}$ as a convolutional kernel. Applying this kernel to the training image yields the inner-products $\langle \vx_{t,\Omega_\ell}, \vu \rangle$ over all spatial locations in a single convolution pass. The quadratic distance is then obtained as $\|\vx_{t,\Omega_\ell}\|^2-2\sqrt{\bar{\alpha}_{t}}\langle \vx_{t,\Omega_\ell}, \vu \rangle+\bar{\alpha}_{t}\|\vu\|^2$.
\looseness=-1

This approach leverages GPU-optimized convolutions to reduce memory usage. We further parallelize across the $L^2$ test patches by batching $B$ patches into a convolutional kernel with $B$ output channels. This yields peak memory $\mathcal{O}(B N L^2)$, which does not explicitly scale with patch size $P$, thereby avoiding the prohibitive $P^2$ factor of naive unfolding and enabling scalable attribution on large datasets.
\looseness=-1

\subsection{Discussions}

Data attribution is often regarded as a lens for understanding model behavior, quantifying the causal influence of individual training examples on model predictions through the learning process. From this perspective, it might seem paradoxical to speak of attribution without specifying a model: if no model is given, what exactly is being attributed?

Our work invites a broader interpretation. In a narrow sense, our nonparametric method can be understood as a principled guess of how training data might have influenced a generated sample, assuming it was produced by some model trained on the same data. Since our method does not access model parameters, it produces the same attribution for all possible models trained on the dataset. At first glance this may appear counterintuitive, because different models can generalize in different ways and might be expected to yield different attribution patterns. However, our empirical results show that the attribution scores remain consistent across a variety of architectures and training regimes. This suggests that there exists a shared, model-agnostic structure to generalization, which reflects intrinsic relationships between training data and outputs, independent of any specific model.

Seen from an even broader perspective, this idea resonates with how humans attribute provenance. People are often able to determine which training images most likely inspired a generated image even without any knowledge of the mechanism that produced it. This form of attribution is grounded in \emph{perceived similarity} rather than \emph{parametric causality}, and it applies whether the image was produced by a neural network, a human artist, or some natural process. In this light, nonparametric data attribution can be viewed as an attempt to formalize this intuitive, model-independent notion of~influence: some training examples are simply more responsible for a given generation than others.

\vspace{-2mm}
\section{Experiments}
\vspace{-1mm}

We evaluate our method, \emph{Nonparametric Diffusion Attribution} (\textbf{NDA}), against both nonparametric and gradient-based attribution approaches using two complementary protocols: the linear datamodeling score (LDS), which quantifies alignment with ground-truth influence, and counterfactual evaluation, which assesses the effect of removing influential training data on generated outputs. Our results and ablations show that NDA achieves strong attribution performance without accessing model parameters or gradients, while qualitative visualizations highlight its spatial interpretability and visual consistency.\footnote{Code is available at \url{https://github.com/sail-sg/NDA}.}
\looseness=-1

\vspace{-2mm}
\subsection{Experimental setup}
\vspace{-1mm}
\textbf{Datasets.}
We conduct experiments on CIFAR-10~\citep{Krizhevsky2009LearningML} and CelebA~\citep{Liu_2015_ICCV}. For efficient ablation studies, we follow \citet{zheng2024intriguing} and construct a CIFAR-2 subset consisting of $5{,}000$ images from CIFAR-10. Additional dataset details are provided in Appendix~\ref{appendix:b1}.

\textbf{Target models.}
For each dataset, we train a diffusion model to serve as the target model for data attribution. On CIFAR-2 and CIFAR-10, we adopt the original DDPM implementation~\citep{ho2020denoising} to train an unconditional diffusion model with a U-Net backbone containing $35.7$M parameters. For CelebA, we use the same implementation but with a modified architecture of $118.8$M parameters to accommodate the $64{\times}64$ resolution. The number of diffusion steps is fixed to $T{=}1{,}000$. Further training details are provided in Appendix~\ref{appendix:b2}. Note that gradient-based attribution methods directly use the target model parameters, whereas our NDA does not access model parameters or architectures.

\textbf{LDS evaluation.}
Following \citet{zheng2024intriguing}, we sample $M{=}64$ random subsets of the training set $\mathbb{S}$, each containing $50\%$ of the samples. For each subset $\mathbb{S}_m$, we train three models with different random seeds and average their $\mathcal{L}_{\text{Simple}}$ losses as the model output $\boldsymbol{\mathcal{F}}(\vx;\theta^{*}(\mathbb{S}_m))$ for a test input $\vx$. Additional implementation details are provided in Appendix~\ref{appendix:b3}. To ensure a fair comparison, we use the same $1{,}000$-image held-out validation set and $1{,}000$-image generation set as \citet{zheng2024intriguing}.

\textbf{NDA setup.}
For patch-wise influence, we use patch sizes $P\in[3,21]$ for different timesteps $t\in\mathcal{T}$ and resolutions, as determined by ablation studies in Sec.~\ref{subsec:abla}. To compute low-resolution patch-wise influence, we apply an average-pooling downsampling operator $\text{D}(\cdot)$ with a window size of $2$, reducing patch resolution by half. In all experiments, we select the top-$k$ most influential patches per training image with $k{=}100$ to obtain image-level attribution scores. Due to low signal-to-noise ratios at large timesteps, we restrict the set of timesteps to $\mathcal{T}{=}\{100,200,300,400,500\}$ in our experiments.
\looseness=-1

\begin{table}[t]
\vspace{-8mm}
\centering
\caption{LDS (\%) of different attribution methods on CIFAR-2, CIFAR-10, and CelebA.}
\label{tab:lds_all}
\vspace{-3mm}
\small
\resizebox{\linewidth}{!}{
\begin{tabular}{lcccccc}
\toprule
\multirow{2}{*}{\textbf{Method}} 
& \multicolumn{2}{c}{\textbf{CIFAR2}} 
& \multicolumn{2}{c}{\textbf{CIFAR10}} 
& \multicolumn{2}{c}{\textbf{CelebA}} \\
\cmidrule(lr){2-3} \cmidrule(lr){4-5} \cmidrule(lr){6-7}
& Validation & Generation & Validation & Generation & Validation & Generation \\
\midrule
\textit{Without Model Access} & & & & & & \\
Raw pixel (dot prod.) & 7.77$\pm$0.57 & 4.89$\pm$0.58 & 2.50$\pm$0.42 & 2.25$\pm$0.39 & 5.58$\pm$0.73 & 4.94$\pm$1.58 \\
Raw pixel (cosine)    & 7.87$\pm$0.57 & 5.44$\pm$0.57 & 2.71$\pm$0.41 & 2.61$\pm$0.38 & 6.16$\pm$0.75 & 4.38$\pm$1.63 \\
CLIP similarity (dot prod.) & 6.51$\pm$1.06 & 3.00$\pm$0.95 & 2.39$\pm$0.41 & 1.11$\pm$0.47 & 8.87$\pm$1.14 & 2.51$\pm$1.13 \\
CLIP similarity (cosine)   & 8.54$\pm$1.01 & 4.01$\pm$0.85 & 3.39$\pm$0.38 & 1.69$\pm$0.49 & 10.92$\pm$0.87 & 3.03$\pm$1.13 \\
\textbf{NDA (Ours)} & \textbf{24.88$\pm$0.42} & \textbf{15.91$\pm$0.49} & \textbf{11.81$\pm$0.30} & \textbf{7.41$\pm$0.45} & \textbf{16.89$\pm$0.59} & \textbf{13.92$\pm$0.68} \\
\midrule
\textit{Using Model Gradients} & & & & & & \\
Gradient (dot prod.) & 5.14$\pm$0.60 & 2.80$\pm$0.55 & 0.79$\pm$0.43 & 0.74$\pm$0.45 & 3.82$\pm$0.50 & 3.83$\pm$1.06 \\
Gradient (cosine)    & 5.08$\pm$0.59 & 2.78$\pm$0.54 & 0.66$\pm$0.43 & 0.58$\pm$0.42 & 3.65$\pm$0.52 & 3.86$\pm$0.96 \\
TracInCP             & 6.26$\pm$0.84 & 3.76$\pm$0.61 & 0.98$\pm$0.44 & 0.96$\pm$0.40 & 5.14$\pm$0.75 & 5.18$\pm$1.05 \\
GAS                  & 5.78$\pm$0.82 & 3.34$\pm$0.56 & 0.89$\pm$0.48 & 0.90$\pm$0.41 & 5.44$\pm$0.68 & 4.69$\pm$0.97 \\
\cmidrule(lr){2-7}
TRAK   & 11.42$\pm$0.49 & 5.78$\pm$0.48 & 2.93$\pm$0.46 & 2.20$\pm$0.38 & 11.28$\pm$0.47 & 7.02$\pm$0.89 \\
D-TRAK & 26.79$\pm$0.33 & 18.82$\pm$0.43 & 14.69$\pm$0.46 & 11.05$\pm$0.43 & 22.83$\pm$0.51 & 16.84$\pm$0.54 \\
\bottomrule
\end{tabular}
}
\vspace{-4mm}
\end{table}

\begin{figure}[tb]
    \centering
    \begin{subfigure}[t]{0.24\textwidth}
        \includegraphics[width=\linewidth]{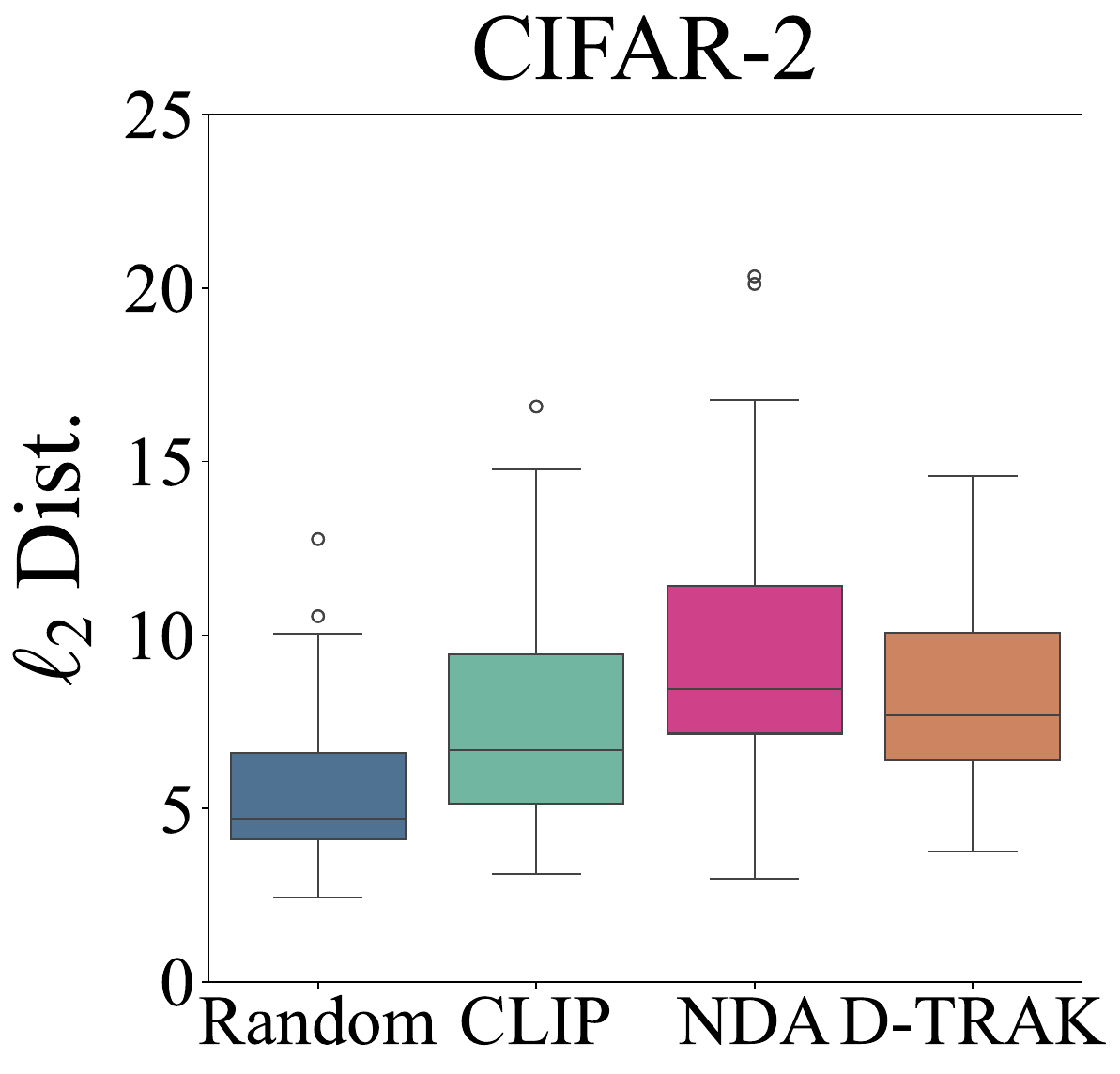}
    \end{subfigure}\hfill
    \begin{subfigure}[t]{0.24\textwidth}
        \includegraphics[width=\linewidth]{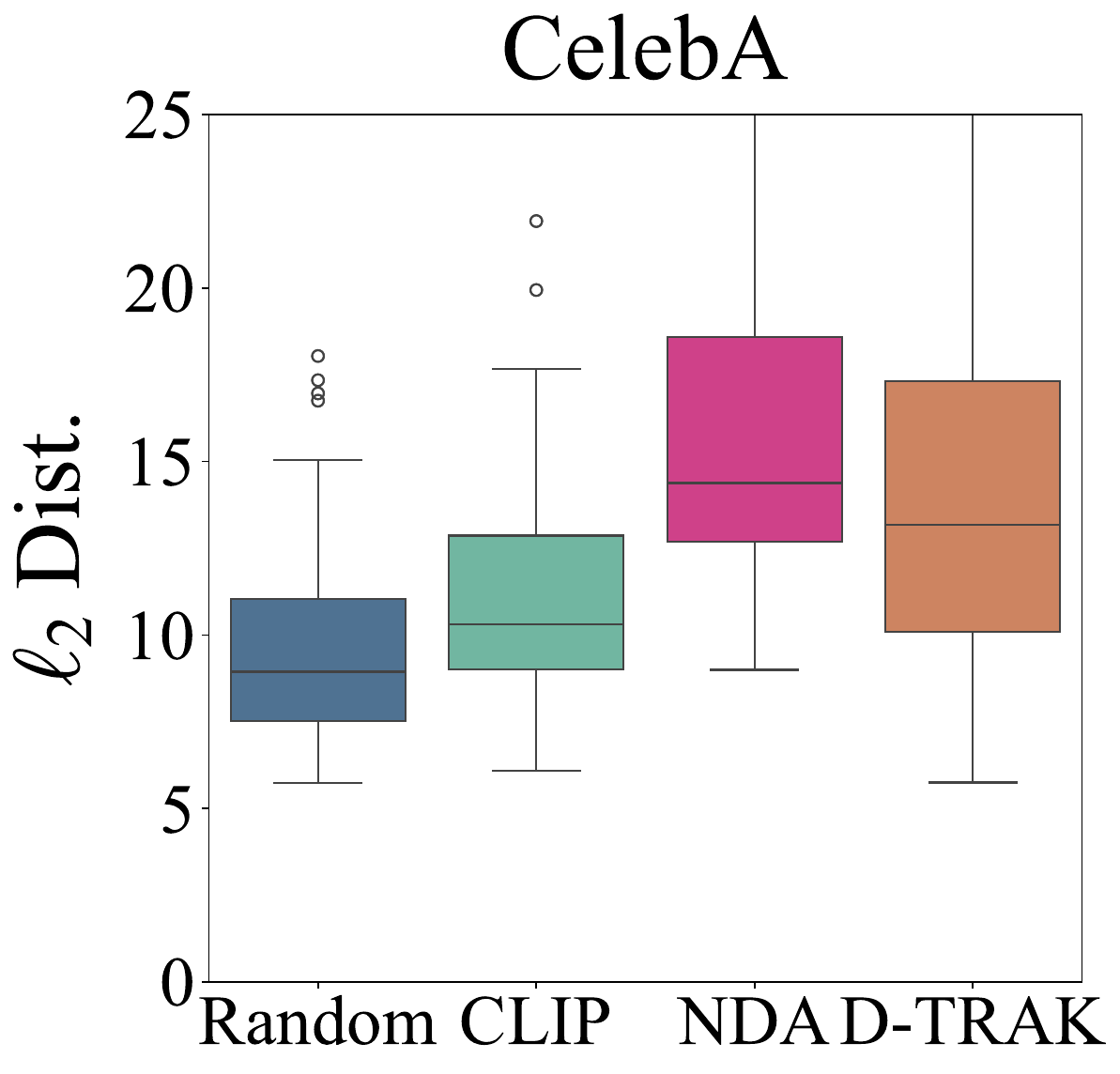}
    \end{subfigure}\hfill
    \begin{subfigure}[t]{0.24\textwidth}
        \includegraphics[width=\linewidth]{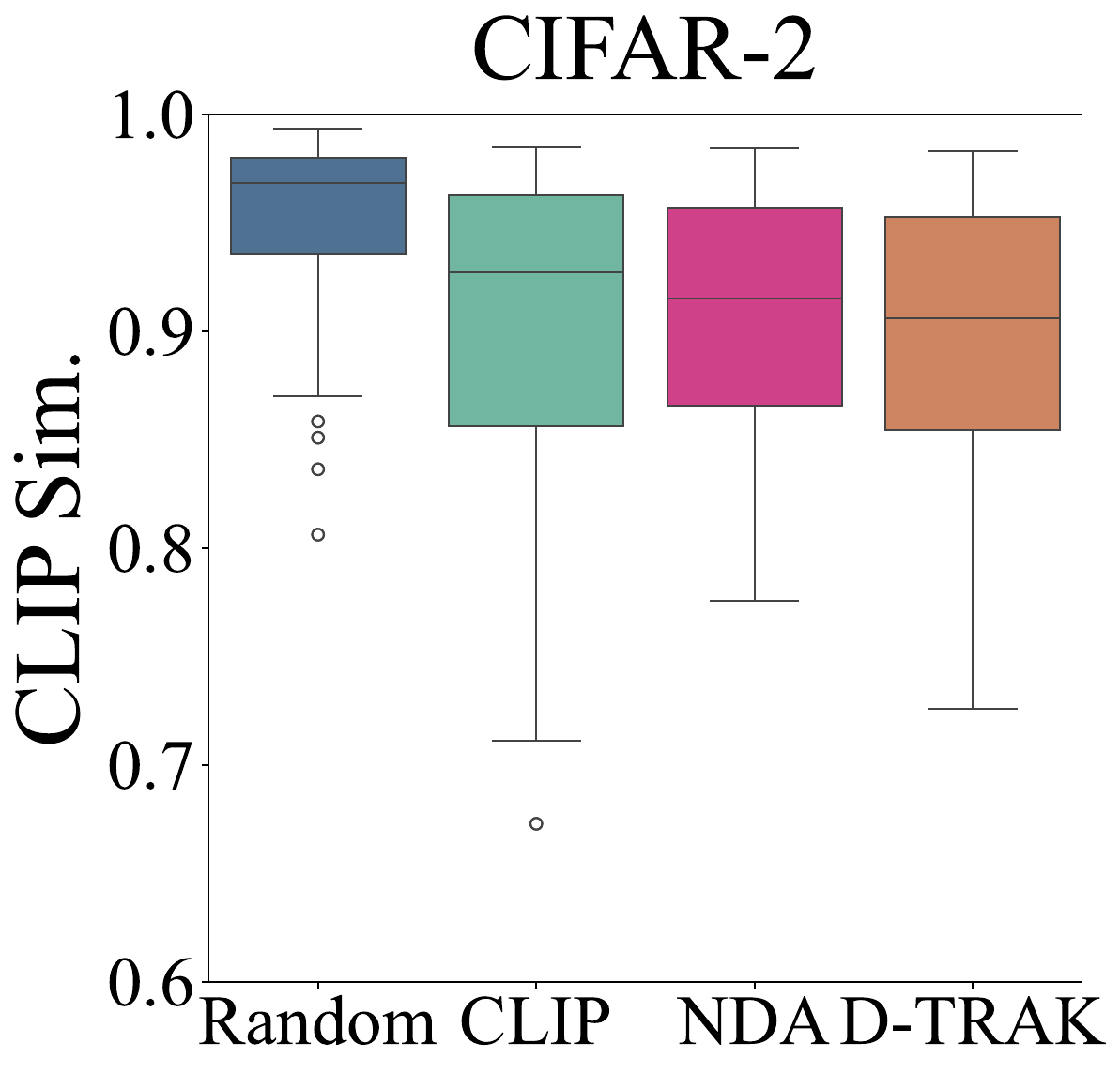}
    \end{subfigure}\hfill
    \begin{subfigure}[t]{0.24\textwidth}
        \includegraphics[width=\linewidth]{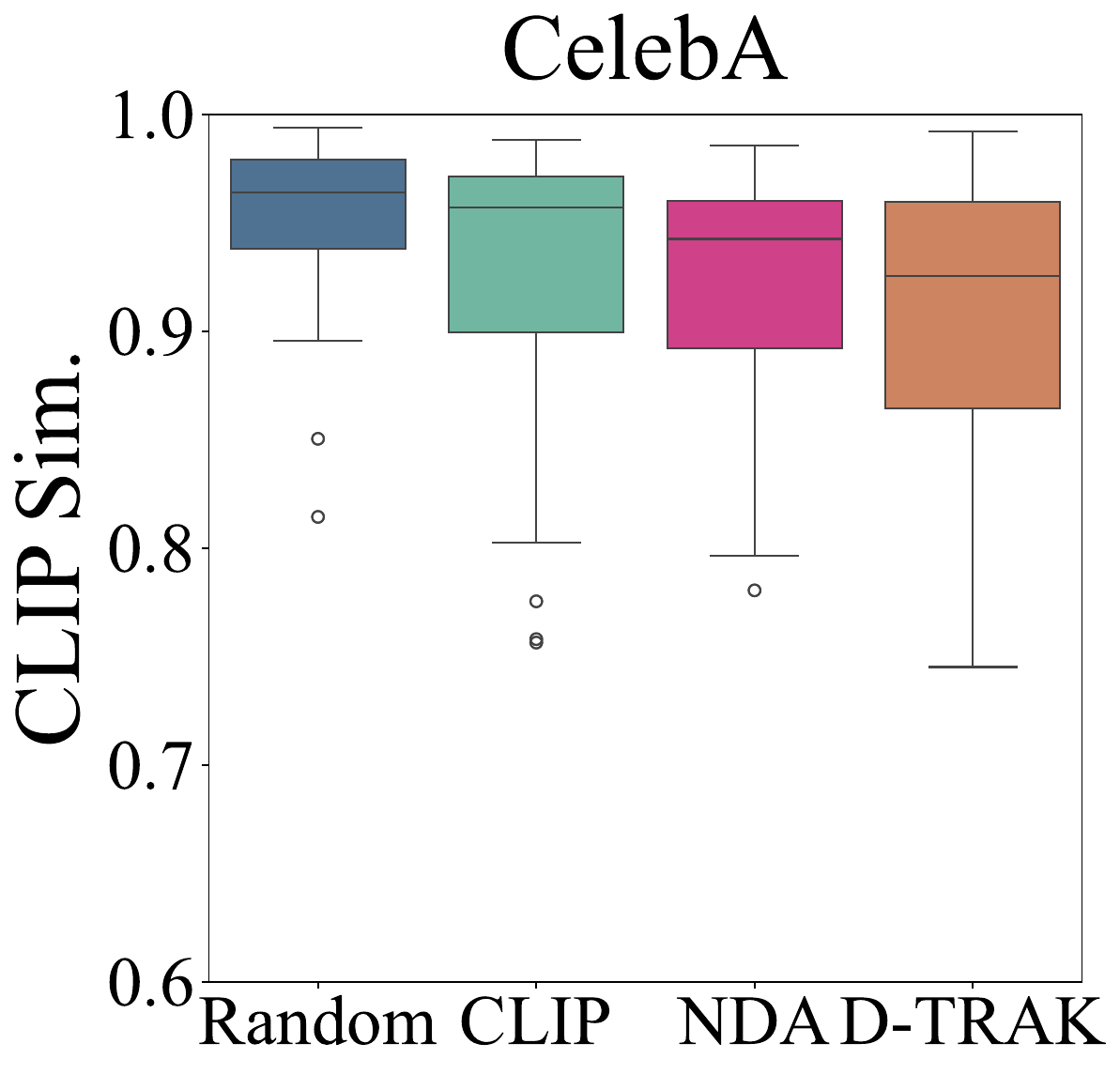}
    \end{subfigure}
    \vspace{-2mm}
    \caption{
    Counterfactual evaluation of $\ell_2$ distance (\textbf{Left}) and CLIP similarity (\textbf{Right}) between original and regenerated images on CIFAR-2 and CelebA after removing the most influential training samples identified by different attribution methods and retraining the model.
    }
    \label{fig:counter_eval}
    \vspace{-4mm}
\end{figure}

\subsection{Main results: evaluating LDS for attribution methods}
\label{subsec:main_results}
We compare NDA against a range of attribution baselines; detailed descriptions are provided in Appendix~\ref{appendix:baselines}. Our primary comparison focuses on representative approaches that do not require access to model parameters, with gradient-based methods included as references. Table~\ref{tab:lds_all} reports the results on CIFAR-2, CIFAR-10, and CelebA.

Compared to the strongest baseline based on CLIP cosine similarity, NDA achieves consistent and substantial gains across both validation and generation sets. On CIFAR-2, NDA improves over CLIP by $+16.32$ (validation) and $+11.90$ (generation); on CIFAR-10, by $+8.42$ and $+5.72$; and on CelebA, by $+5.97$ and $+10.89$, respectively.
When compared to strong gradient-based methods tailored for diffusion models, such as D-TRAK, NDA substantially closes the gap while accessing no model information. For example, on CelebA, the gap is reduced to $5.94$/$2.92$ (validation/generation), much smaller than the CLIP gap of $11.91$/$13.81$.
Overall, NDA approaches the performance of strong parametric baselines while markedly outperforming CLIP similarity on LDS.

\vspace{-1mm}
\subsection{Counterfactual evaluation}
\vspace{-1mm}
To evaluate the faithfulness and practical effectiveness of NDA, we conduct a counterfactual influence experiment on CIFAR-2 and CelebA.
For each generated test image, we first identify the top-$1{,}000$ most positively influential training samples according to each attribution method, remove these samples from the training set, and retrain the diffusion model from scratch. The test image is then regenerated using the retrained models under the same random seed, and the impact of removal is quantified using pixel-wise $\ell_2$ distance and CLIP cosine similarity. We repeat this procedure for $60$ randomly generated test images and report the average results.

\begin{figure}[t]
    \centering
    \vspace{-8mm}
    \scriptsize
    \resizebox{\linewidth}{!}{
    \begin{tabular}{c@{\hspace{2em}}c}
        \begin{tabular}{@{}c@{}c@{}c@{}c@{}c@{}}
             & \textbf{Random} & \textbf{CLIP} & \textbf{NDA (Ours)} & \textbf{D-TRAK} \\
            \includegraphics[width=0.1\textwidth]{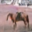} &
            \includegraphics[width=0.1\textwidth]{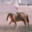} &
            \includegraphics[width=0.1\textwidth]{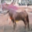} &
            \includegraphics[width=0.1\textwidth]{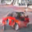} &
            \includegraphics[width=0.1\textwidth]{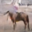} \\
            \includegraphics[width=0.1\textwidth]{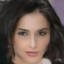} &
            \includegraphics[width=0.1\textwidth]{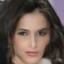} &
            \includegraphics[width=0.1\textwidth]{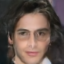} &
            \includegraphics[width=0.1\textwidth]{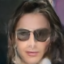} &
            \includegraphics[width=0.1\textwidth]{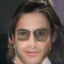}
        \end{tabular}
        &
        \begin{tabular}{@{}c@{}c@{}c@{}c@{}c@{}}
             & \textbf{Random} & \textbf{CLIP} & \textbf{NDA (Ours)} & \textbf{D-TRAK} \\
            \includegraphics[width=0.1\textwidth]{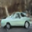} &
            \includegraphics[width=0.1\textwidth]{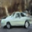} &
            \includegraphics[width=0.1\textwidth]{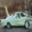} &
            \includegraphics[width=0.1\textwidth]{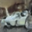} &
            \includegraphics[width=0.1\textwidth]{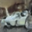} \\
            \includegraphics[width=0.1\textwidth]{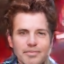} &
            \includegraphics[width=0.1\textwidth]{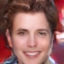} &
            \includegraphics[width=0.1\textwidth]{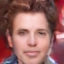} &
            \includegraphics[width=0.1\textwidth]{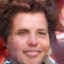} &
            \includegraphics[width=0.1\textwidth]{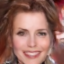}
        \end{tabular}
    \end{tabular}
    }
    \vspace{-3mm}
    \caption{Counterfactual visualization on CIFAR-2 (\textbf{Top}) and CelebA (\textbf{Bottom}). Images are compared to those generated by retrained models using the same seed. See Appendix~\ref{appendix:vis_counter} for more cases.\looseness=-1}
    \label{fig:counter_vis1}
    \vspace{-4mm}
\end{figure}

We compare NDA against three baselines: Random removal, CLIP similarity (cosine), and D-TRAK. As shown in Figure~\ref{fig:counter_eval}, NDA achieves median $\ell_2$ distances of $8.43$ and $14.38$ on CIFAR-2 and CelebA, respectively, outperforming CLIP ($6.68, 10.30$) and D-TRAK ($7.68, 13.18$). For CLIP similarity, NDA attains $0.92$ and $0.94$, approaching D-TRAK ($0.91, 0.93$). These results indicate that NDA effectively identifies and removes training samples with strong influence over the generated output, providing a compelling nonparametric alternative to gradient-based attribution. Qualitative visualizations in Figure~\ref{fig:counter_vis1} further show that the generated images exhibit significant distortions after removing the training samples deemed most influential by NDA.

\subsection{Spatial interpretability}

\begin{figure}[t]
    \centering
    \includegraphics[width=.97\textwidth]{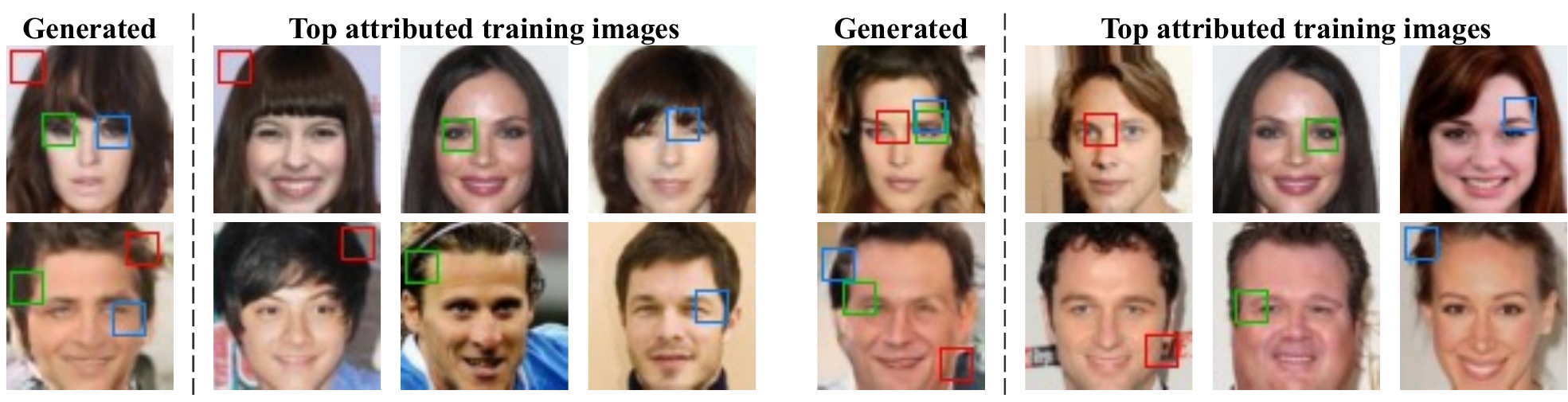}
    \vspace{-3mm}
    \caption{Spatial interpretability of NDA. The patches with the highest patch-wise influence scores (w.r.t.\ patches in the generated image) are highlighted in the top attributed training images.\looseness=-1}
\label{fig:spatial_interp}
\vspace{-3mm}
\end{figure}

Since NDA computes attribution by aggregating patch-wise influence, it naturally offers an additional spatial level of interpretability. Intuitively, a training image with a high attribution score must contain patches that strongly align with influential regions of the test image. To visualize this correspondence, we highlight the most influential patches from each of the top attributed training images in Figure~\ref{fig:spatial_interp}.
We observe that top-ranked training images consistently contain local patches that are visually similar to those in the test image, offering a natural explanation for their high attribution scores.

\begin{figure}[!t]
    \centering
    \begin{subfigure}[t]{0.32\textwidth}
        \centering
        \includegraphics[width=\linewidth]{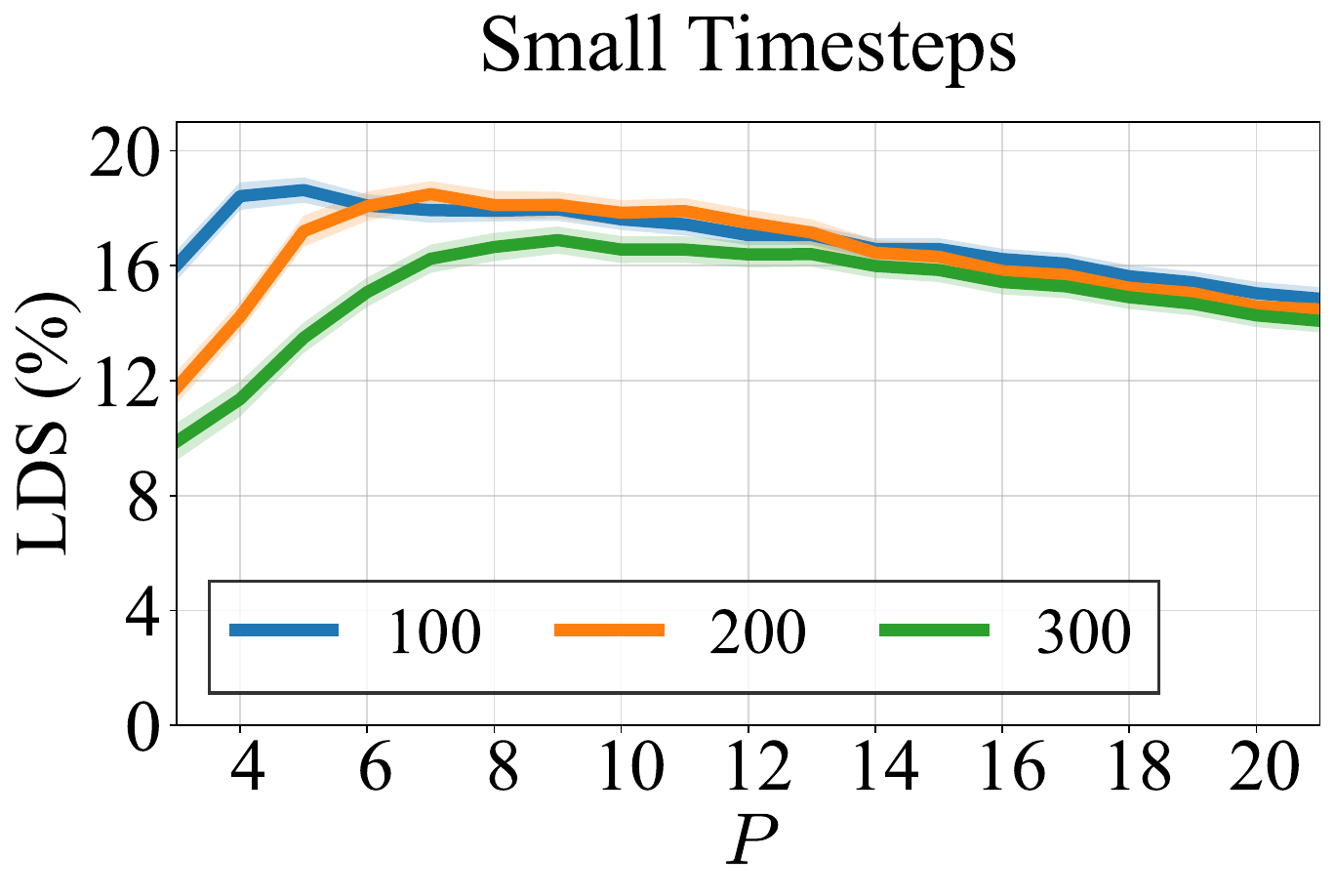}
    \end{subfigure}
    \hfill
    \begin{subfigure}[t]{0.32\textwidth}
        \centering
        \includegraphics[width=\linewidth]{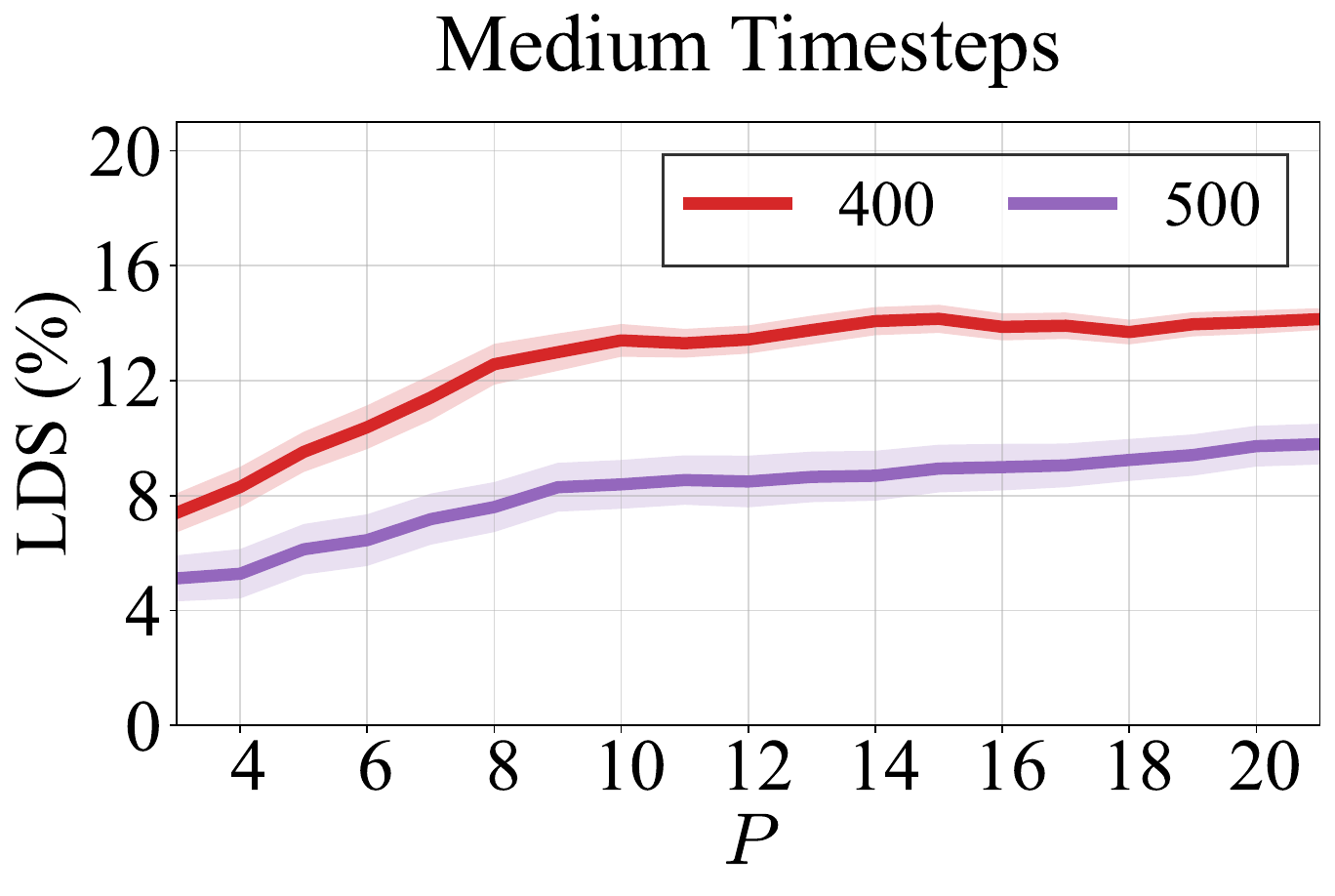}
    \end{subfigure}
    \hfill
    \begin{subfigure}[t]{0.32\textwidth}
        \centering
        \includegraphics[width=\linewidth]{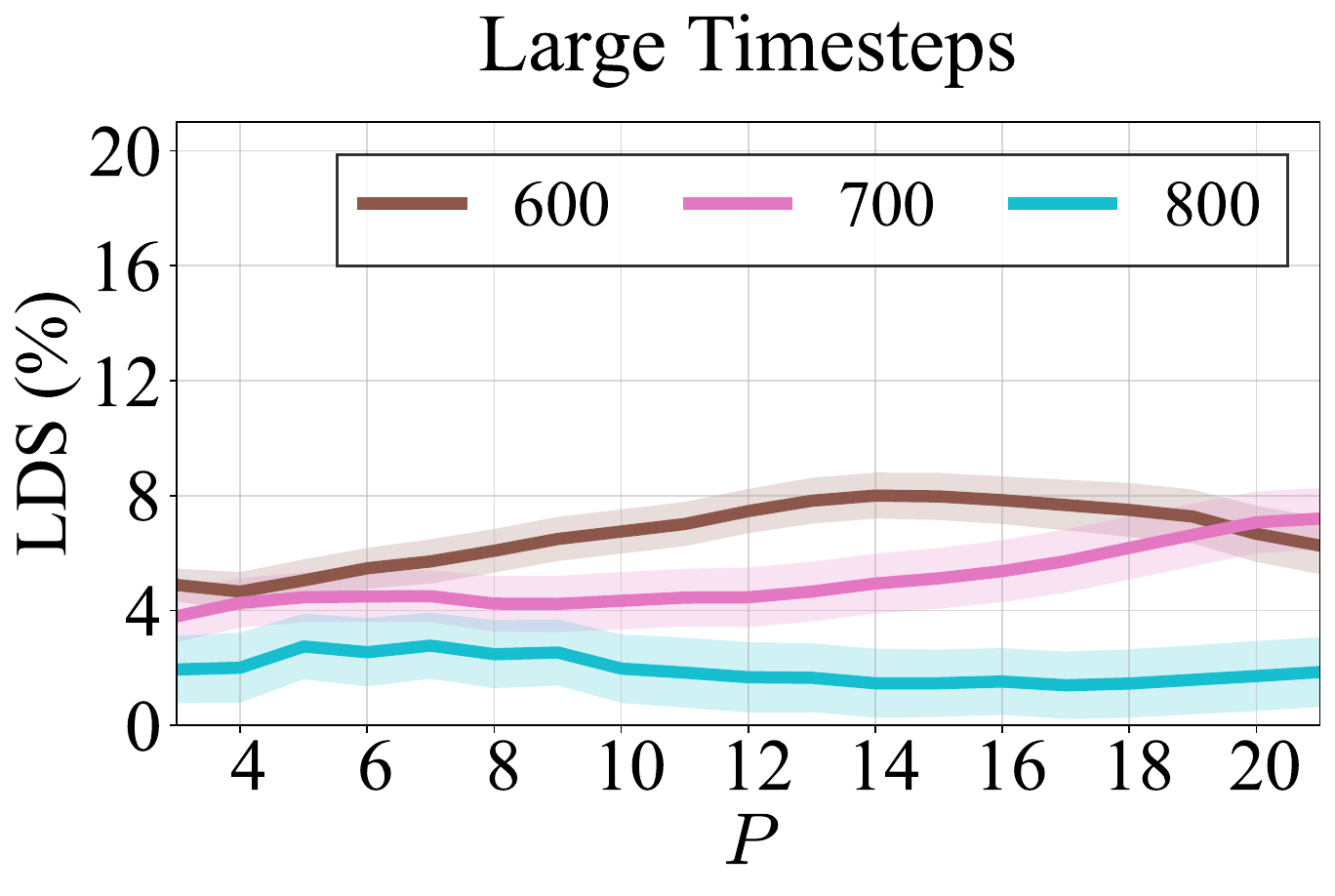}
    \end{subfigure}
    \vspace{-3mm}
    \caption{
        LDS (\%) on the CIFAR-2 validation set across different patch sizes and timesteps at the original resolution. \textbf{Left}: small timesteps ($t{\leq}300$), where moderate patch sizes achieve the highest scores. \textbf{Middle}: medium timesteps ($t{=}400,500$), where larger patches outperform smaller ones due to higher noise. \textbf{Right}: large timesteps, where noise dominates and informative signal is minimal.\looseness=-1
    }
\label{fig:cifar2_ps-vs-timestep}
\vspace{-4mm}
\end{figure}

\vspace{-1mm}
\subsection{Ablation studies}
\vspace{-1mm}
\label{subsec:abla}
We conduct ablation studies on key hyperparameters of our method. All studies are performed on the validation sets, which are i.i.d.\ samples drawn from the same distribution as the training sets. The selected hyperparameters are then applied to the generation sets without further tuning to produce the results reported in Sec.~\ref{subsec:main_results}.
In all figures where applicable, solid lines show the mean, and shaded regions indicate $\pm 1$ standard deviation.

\textbf{Patch size selection.}
We study the effect of patch size by evaluating LDS performance across $P\in[3,21]$ at different timesteps. To isolate the impact of timestep, we fix $\mathcal{T}$ to contain a single timestep (e.g., $\mathcal{T}{=}\{100\},\dots,\{900\}$) and vary $P$. Figure~\ref{fig:cifar2_ps-vs-timestep} shows LDS curves as a function of $P$ for each $\mathcal{T}$ on CIFAR-2 (see Appendix~\ref{appendix:abla} for CelebA).
We observe that the optimal patch size varies with timestep and generally increases as $t$ grows. At early timesteps ($t{\leq}300$), small to moderate patches ($P{=}5,7,9$) yield the highest LDS scores, suggesting that local patterns dominate when noise levels are relatively low during reverse diffusion. For mid-range timesteps ($400{\leq}t{\leq}500$), larger patches perform better, due to their ability to aggregate more contextual information under higher noise levels. In the high-noise regime ($t{\geq}600$), LDS scores drop significantly across all patch sizes. Based on these results, we adopt a timestep-dependent patch size selection strategy: using smaller patches for early timesteps and progressively larger patches for later timesteps, choosing the $P$ that maximizes LDS for each $t$.
\looseness=-1

\begin{figure}[t]
    \vspace{-9mm}
    \centering
    \begin{subfigure}[t]{0.32\textwidth}
        \centering
        \includegraphics[width=\linewidth]{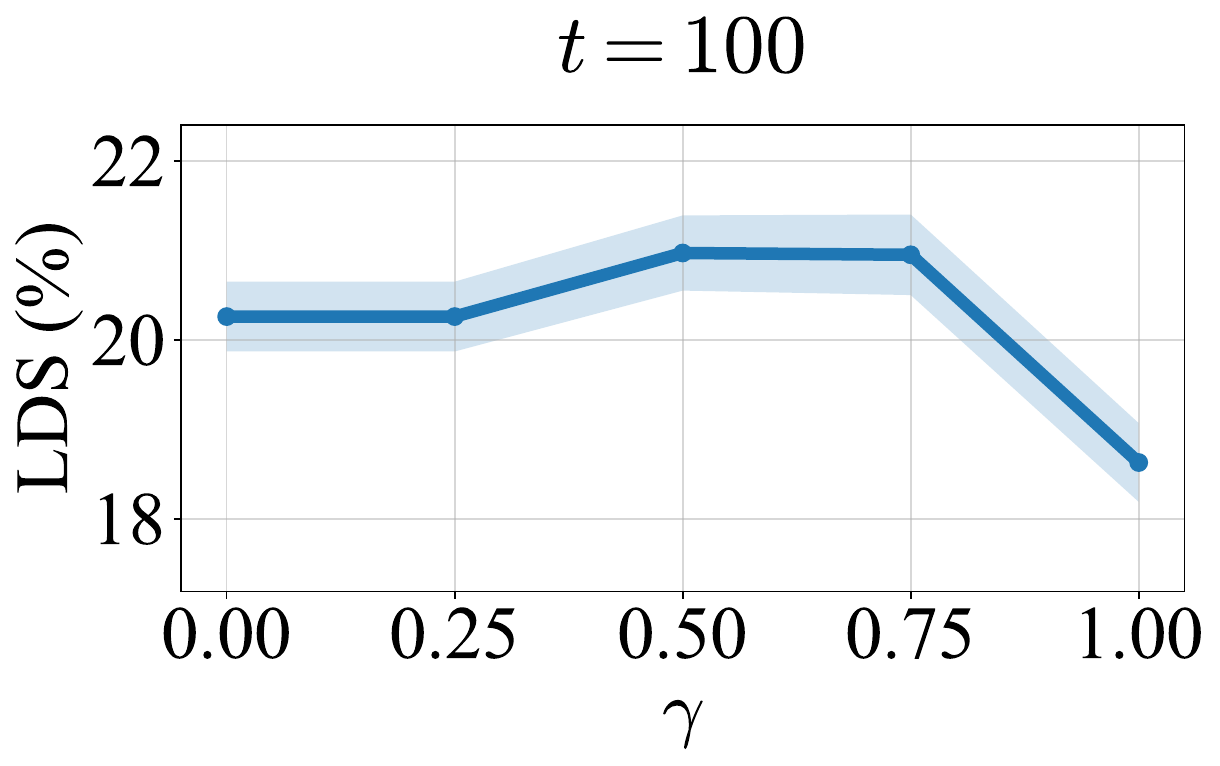}
    \end{subfigure}
    \hfill
    \begin{subfigure}[t]{0.32\textwidth}
        \centering
        \includegraphics[width=\linewidth]{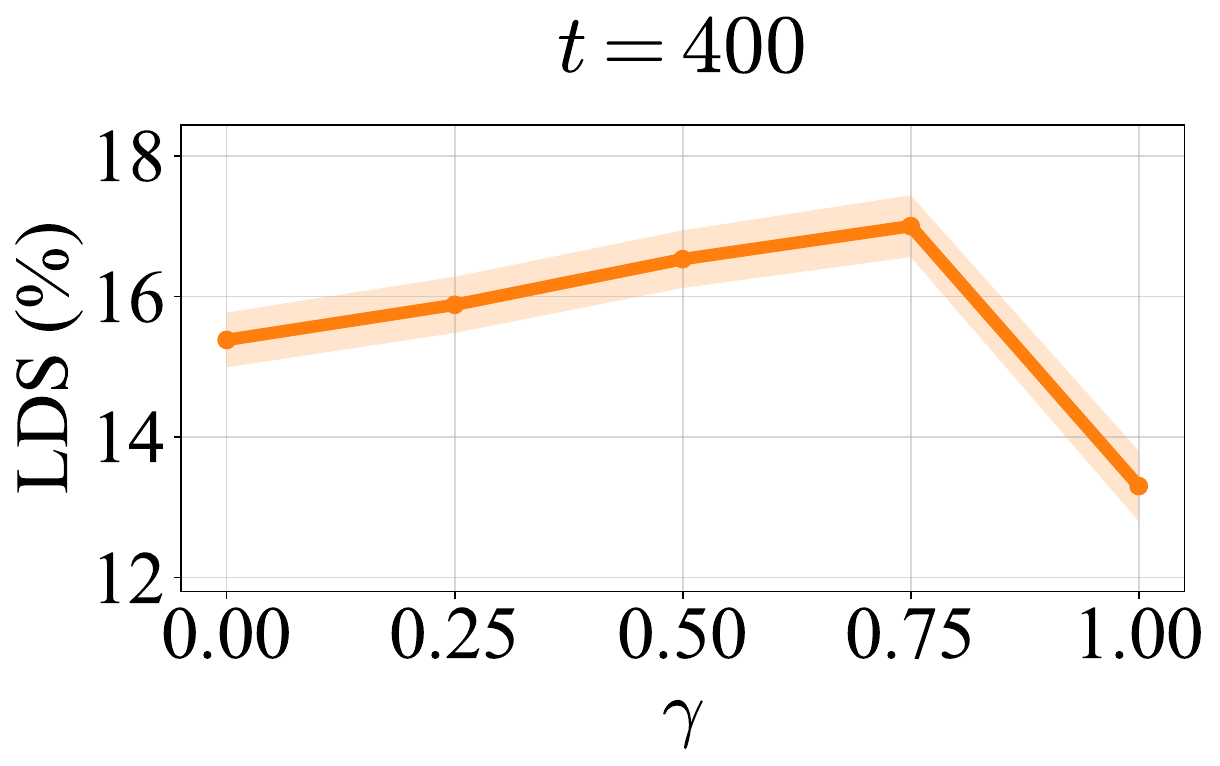}
    \end{subfigure}
    \hfill
    \begin{subfigure}[t]{0.32\textwidth}
        \centering
        \includegraphics[width=\linewidth]{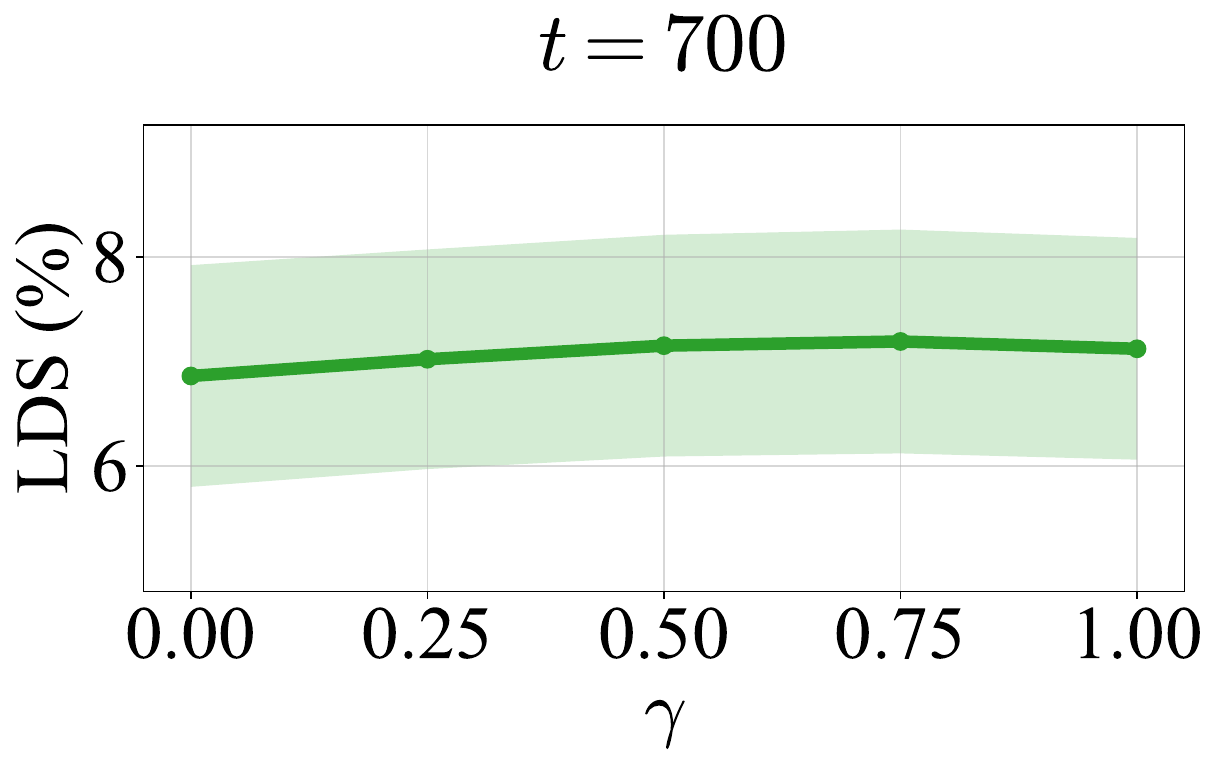}
    \end{subfigure}
    \vspace{-3mm}
    \caption{LDS (\%) on CIFAR-2 across different weighting factor $\gamma$ at varying timesteps.
    }
    \label{fig:cifar2_gamma_multiscale}
    \vspace{2mm}
  \begin{subfigure}[t]{0.24\textwidth}
    \includegraphics[width=\linewidth]{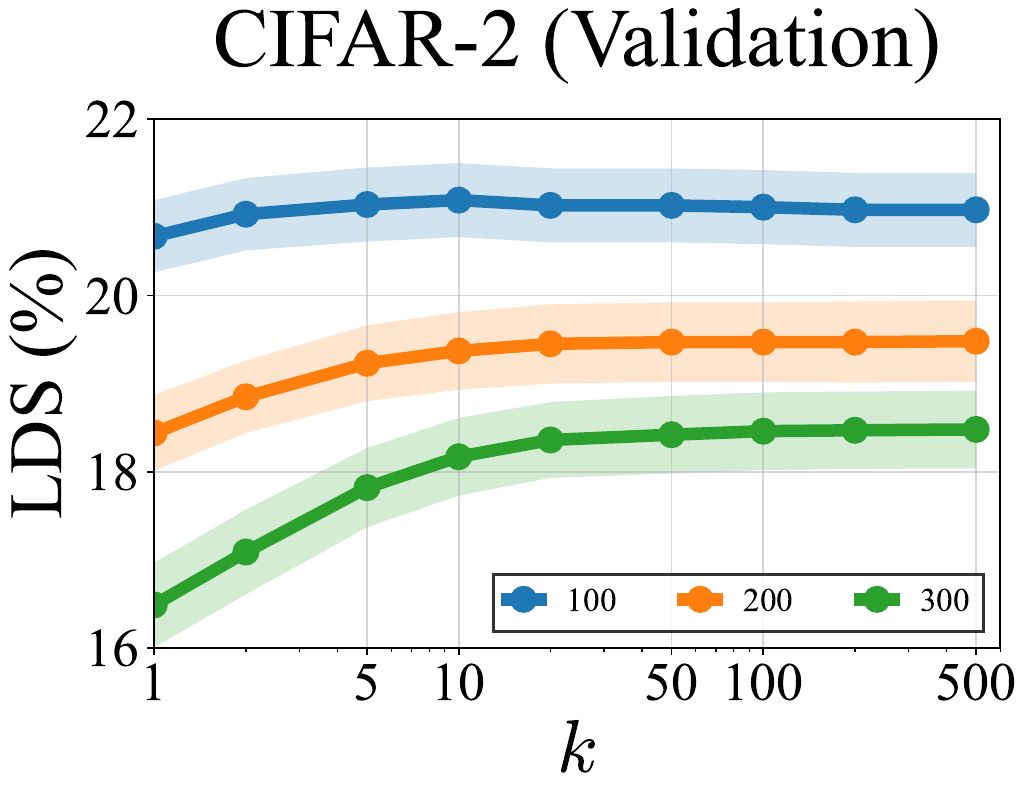}
  \end{subfigure}\hfill
  \begin{subfigure}[t]{0.24\textwidth}
    \includegraphics[width=\linewidth]{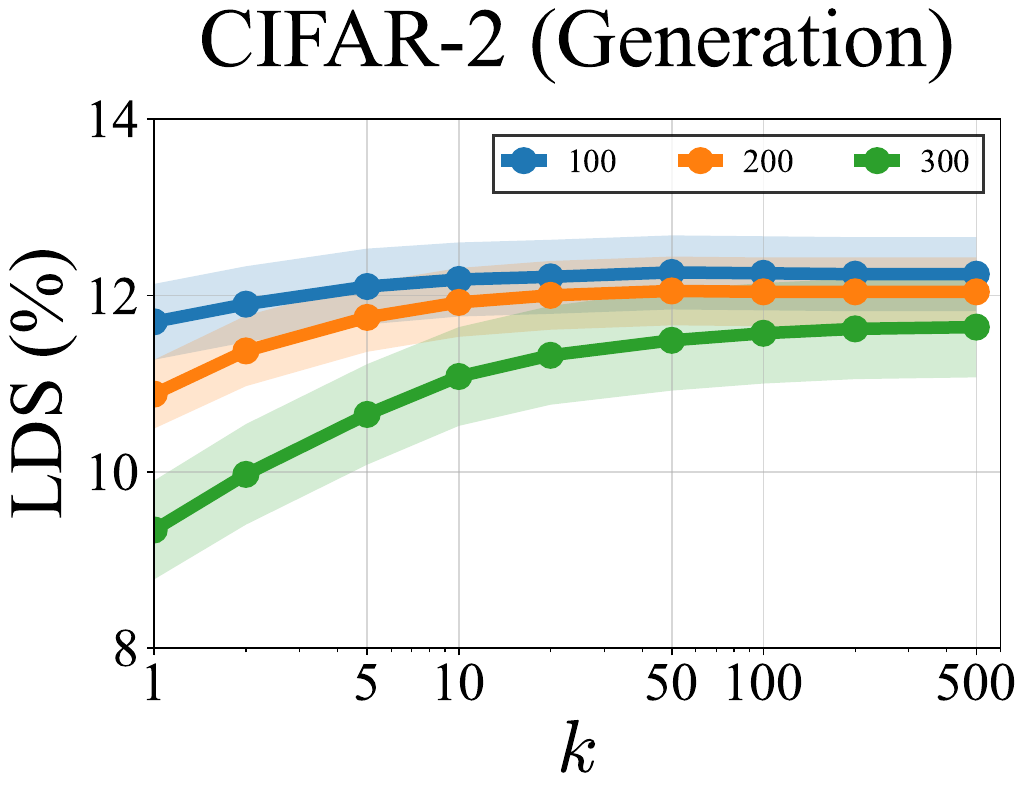}
  \end{subfigure}\hfill
  \begin{subfigure}[t]{0.24\textwidth}
    \includegraphics[width=\linewidth]{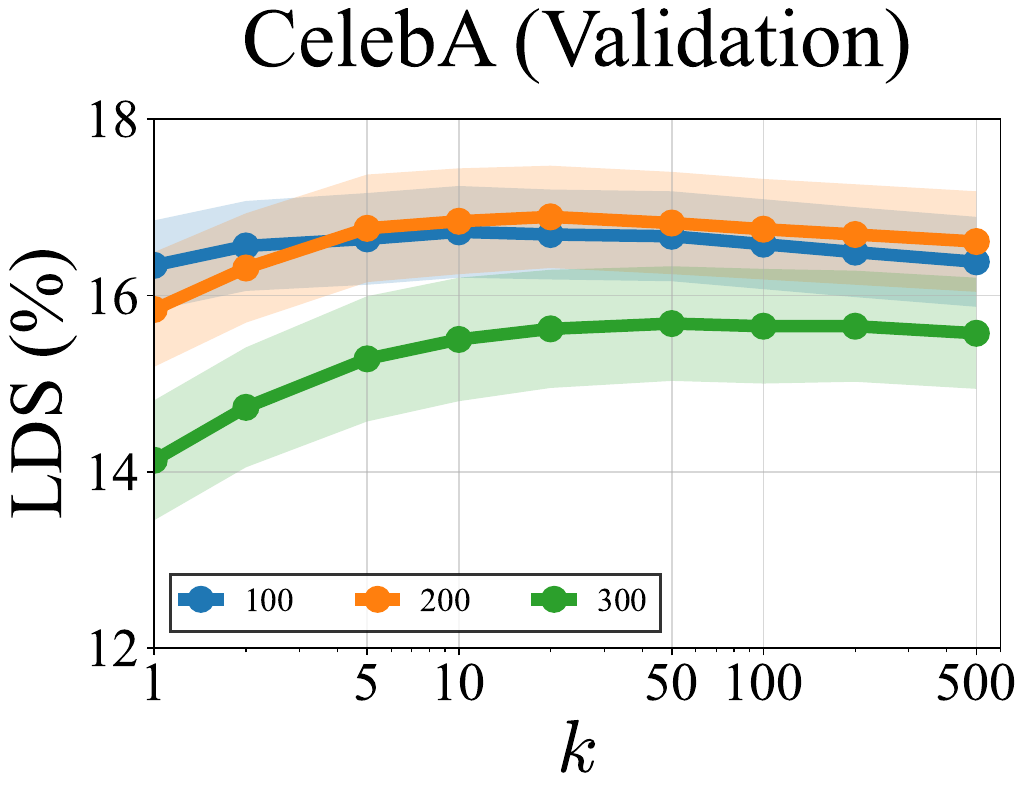}
  \end{subfigure}\hfill
  \begin{subfigure}[t]{0.24\textwidth}
    \includegraphics[width=\linewidth]{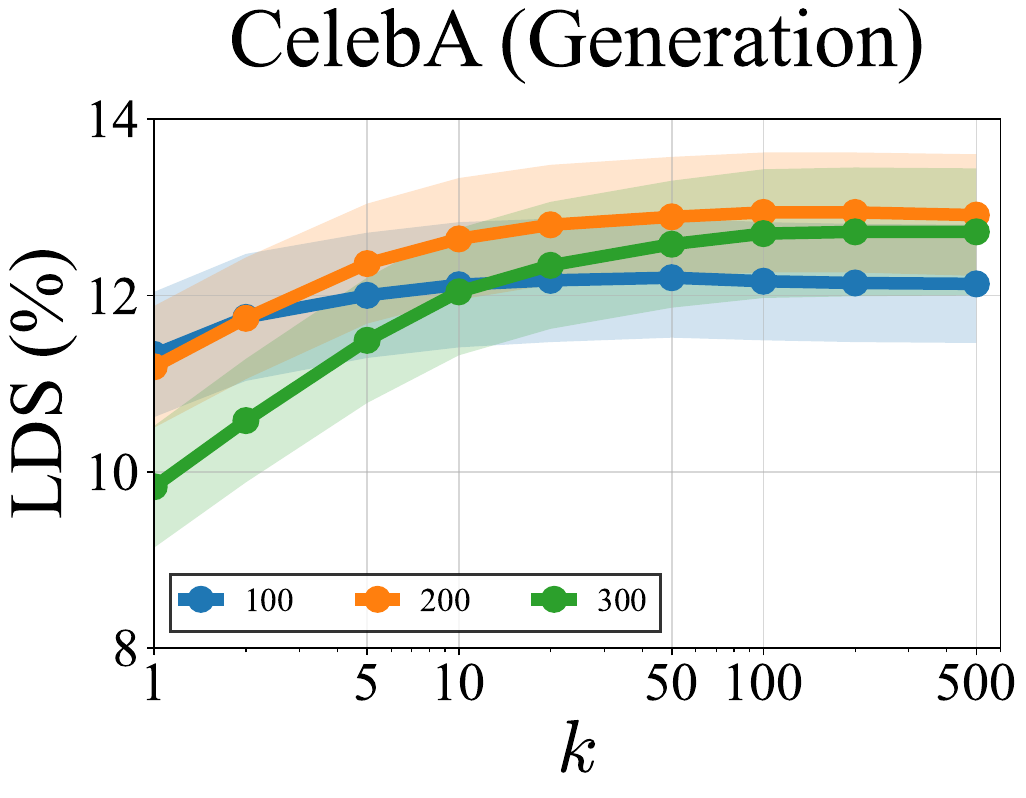}
  \end{subfigure}
  \vspace{-3mm}
  \caption{
    LDS (\%) as a function of the number $k$ of top influential patches selected for aggregation on CIFAR-2 (\textbf{Left}) and CelebA (\textbf{Right}). Setting $k{=}100$ generally provides strong performance.
  }
  \label{fig:topk_selection}
  \vspace{2mm}
  \begin{subfigure}[t]{0.24\textwidth}
    \includegraphics[width=\linewidth]{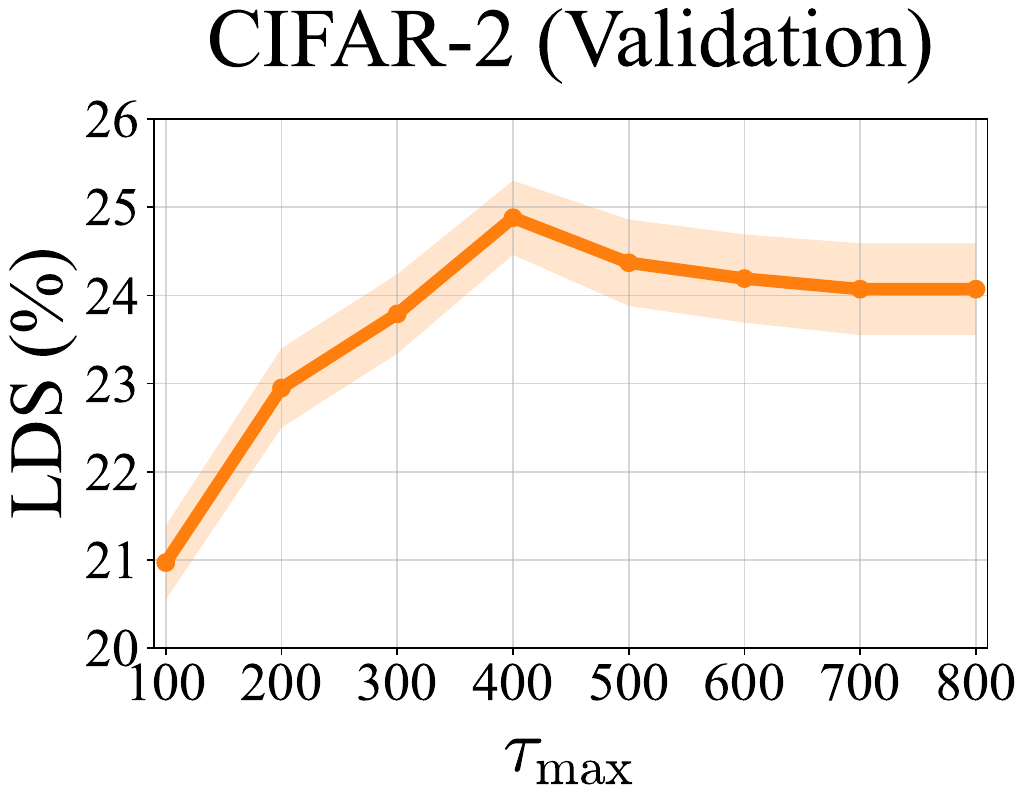}
  \end{subfigure}\hfill
  \begin{subfigure}[t]{0.24\textwidth}
    \includegraphics[width=\linewidth]{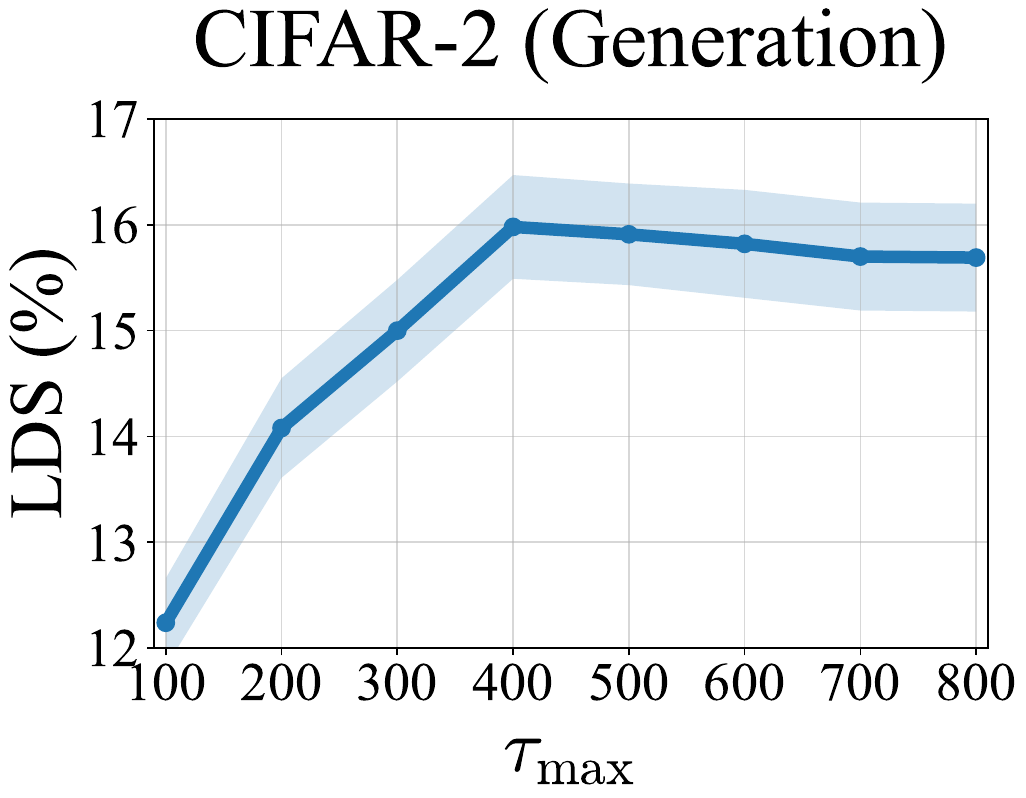}
  \end{subfigure}\hfill
  \begin{subfigure}[t]{0.24\textwidth}
    \includegraphics[width=\linewidth]{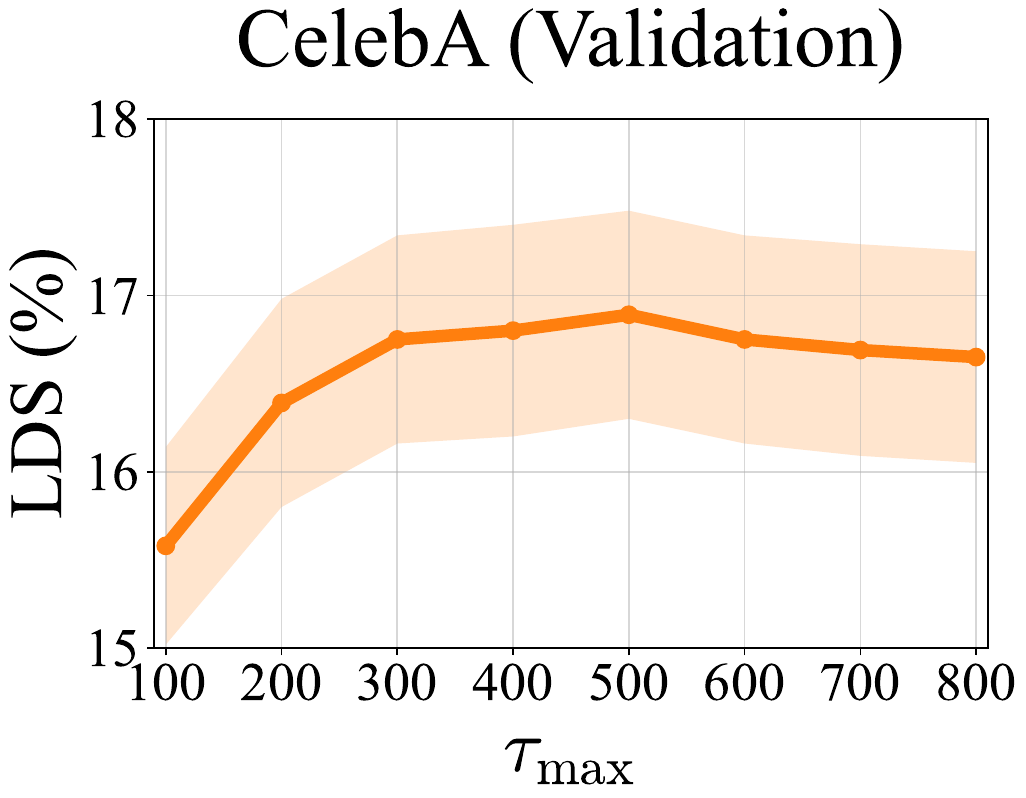}
  \end{subfigure}\hfill
  \begin{subfigure}[t]{0.24\textwidth}
    \includegraphics[width=\linewidth]{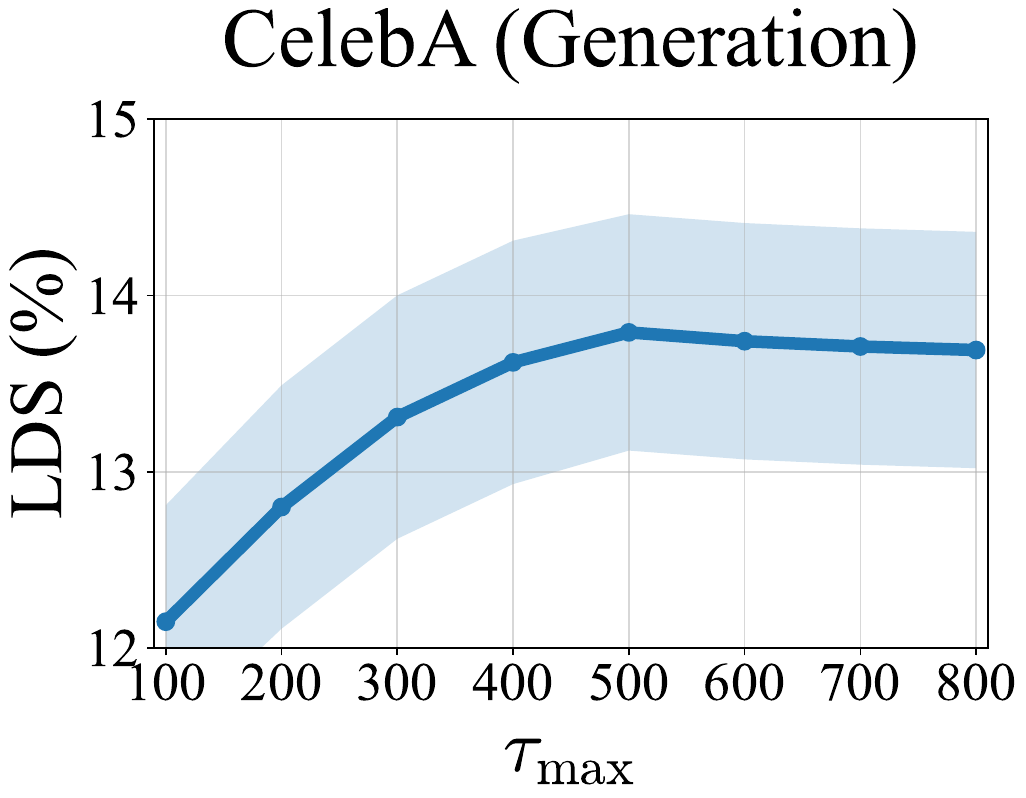}
    \end{subfigure}
    \vspace{-3mm}
  \caption{
    LDS (\%) as a function of the max aggregation timestep $\tau_{\max}$ on CIFAR-2 (\textbf{Left}) and CelebA (\textbf{Right}). Choosing $\tau_{\max}{=}500$ generally yields good results.
  }
  \label{fig:max_averaged_timestep}
  \vspace{-4mm}
\end{figure}

\textbf{Multiscale influence.}
We first find the optimal patch size for the low-resolution patch-wise influence in Eq.~(\ref{eq:patch_softmax_down}) using the same procedure as above and observe a similar trend, where early timesteps favor smaller patches. Interestingly, the optimal patch size differs between the original and low-resolution cases (see Appendix~\ref{appendix:abla} for details).
We then evaluate the proposed multiscale attribution in Eq.~(\ref{eq:image_ms}) by varying the weighting factor $\gamma \in \{0,0.25,0.5,0.75,1\}$ across timesteps. As shown in Figure~\ref{fig:cifar2_gamma_multiscale}, combining multiple scales (e.g., $\gamma{=}0.75$) consistently improves LDS, particularly in the low-noise regime, suggesting that multiscale features provide complementary information that enhances attribution. Further improvements may be possible by incorporating more scales, which we leave for future work.
\looseness=-1

\textbf{Top-$k$ patch selection.}
We examine how the number of top-$k$ patches affects attribution. Figure~\ref{fig:topk_selection} shows LDS as a function of $k$ on lower-noise timesteps, which contribute more significantly to the final attribution. We test $k$ ranging from $1$ to $500$ and find that an intermediate value of $k{=}100$ generally achieves optimal performance across timesteps and datasets and generalizes well to generation sets.

\textbf{Timestep selection.}
Since lower-noise timesteps provide stronger attribution signal, we define $\mathcal{T}=\{100,200,\ldots,\tau_{\max}\}$ and study the effect of varying the max aggregation timestep $\tau_{\max}$. At each $t \in \mathcal{T}$ we use the optimal patch sizes and weighting factors identified above. Figure~\ref{fig:max_averaged_timestep} shows that enlarging $\mathcal{T}$ improves LDS up to mid-range timesteps, after which gains saturate or degrade. On CIFAR-2, performance rises from $t{=}100$ to a peak near $\tau_{\max}{=}400$ before dropping by ${\sim}0.5$ at $\tau_{\max}{\geq}600$; On CelebA, performance saturates near $\tau_{\max}{=}400{\sim}500$ with negligible gains beyond. These results suggest that late (high-noise) timesteps contribute little useful signal and may instead inject noise. We therefore fix $\tau_{\max}{=}500$ for all subsequent experiments, which achieves near-optimal performance while avoiding unnecessary computation and variability.

\vspace{-2mm}
\section{Conclusion}
\vspace{-1mm}
We present a nonparametric method for data attribution in diffusion models that measures the influence of training examples via patch-level similarity. Our approach leverages the analytical form of the optimal score function, extends to multiscale representations, and remains efficient through convolution-based operations. It provides spatially interpretable attributions and uncovers intrinsic relationships between training data and generated outputs, independent of any specific model. Experiments on CIFAR-2, CIFAR-10, and CelebA show that our method closely matches gradient-based approaches and outperforms existing nonparametric baselines, demonstrating that model-agnostic, data-driven attribution is a practical tool for understanding generative models.

\bibliography{ref}
\bibliographystyle{iclr2026_conference}

\clearpage
\appendix

\section{Related works}
\label{appendix:related}
\subsection{Data attribution}
Training data plays a critical role in shaping the behavior of machine learning models. Data attribution aims to measure the contribution of individual training samples to model predictions. Existing methods can be broadly categorized into two classes: retraining-based and retraining-free approaches.

Retraining-based methods, such as empirical influence functions~\citep{NEURIPS2020_1e14bfe2}, Shapley value estimators~\citep{Jia_2021_CVPR}, and datamodels~\citep{ilyas2022datamodels}, provide high-fidelity attributions by measuring the effect of removing or modifying each training example. However, these methods typically require retraining the model tens of thousands of times on different data subsets to achieve reliable results, making them computationally infeasible for large-scale settings.
Retraining-free methods aim to approximate influence scores without additional retraining, offering a more scalable alternative. These approaches fall into two main categories. The first comprises gradient-based methods without kernelization, which rely solely on first-order gradient signals and avoid computing second-order derivatives. Representative approaches include Gradient~\citep{NEURIPS2019_c61f571d}, TracInCP~\citep{NEURIPS2020_e6385d39} and GAS~\citep{10.1145/3548606.3559335}.
The second category includes gradient-based methods with kernelization, which incorporate curvature information by constructing kernels, typically using the inverse of the Hessian matrix.
However, Hessian inversion is often numerically unstable. 
To mitigate this, recent work approximates the Hessian with the Fisher information matrix (FIM). For instance, TRAK~\citep{pmlr-v202-park23c} introduces a kernel-based approximation that efficiently estimates influence scores via random projections, which is both accurate and computationally tractable for large-scale models. D-TRAK~\citep{zheng2024intriguing} further improves attribution performance by modifying the output function and training loss in TRAK, resulting in more effective LDS scores. Based on empirical analysis, D-TRAK recommends a specific configuration that achieves the best overall performance.
\looseness=-1

\vspace{-2mm}
\subsection{Nonparametric data attribution methods}
\vspace{-1mm}
While most data attribution methods rely on parametric models and require gradient computations or model retraining, an alternative line of research explores nonparametric approaches that operate directly on training data without fitting parametric functions. These are typically similarity-based approaches, which estimate the influence of a training sample based on its similarity to the target using a predefined metric.
For example, one can leverage pretrained vision-language embeddings such as CLIP~\citep{radford2021learning} features to measure high-level semantic similarity of images~\citep{zheng2024intriguing, mlodozeniec2025influence}, or compare raw pixel values directly. However, these methods overlook the internal dynamics of the generative model. As a result, effective data attribution for diffusion models in settings without model access remains an open and challenging problem.

\section{Implementation details}
\subsection{Datasets}
\label{appendix:b1}
\textbf{CIFAR-10 ($32{\times}32$).} CIFAR-10 contains $50{,}000$ training and $10{,}000$ test images. For LDS evaluation, we randomly sample $1{,}000$ validation images from the test set. To reduce computational overhead, we additionally construct \textbf{CIFAR-2} ($32{\times}32$), a subset consisting of $5{,}000$ training images randomly selected from the ``automobile'' and ``horse'' classes and $1{,}000$ validation images from the corresponding classes in the test set.

\textbf{CelebA ($64{\times}64$).} We sample $5{,}000$ training and $1{,}000$ validation images from the official CelebA splits~\citep{Liu_2015_ICCV}. Following the preprocessing steps outlined by \citet{song2021scorebased}, we center-crop all images to $140{\times}140$ pixels and resize them to $64{\times}64$.

\subsection{Models}
\label{appendix:b2}
\textbf{CIFAR-10 and CIFAR-2.} We adopt the original unconditional DDPM implementation~\citep{ho2020denoising}, using a U-Net backbone with approximately $35.7$M parameters. Models are trained for $200$ epochs with a batch size of $128$ and cosine annealing learning rate schedule. We use AdamW~\citep{loshchilov2019decoupledweightdecayregularization} with weight decay $10^{-6}$ and a dropout rate of $0.1$. To enhance robustness, random horizontal flips are applied for data augmentation.

\textbf{CelebA.} We use the same DDPM framework but scale the U-Net to $118.8$M parameters to accommodate $64{\times}64$ inputs. All other hyperparameters, including the variance schedule, optimizer configurations, and training procedure, follow the CIFAR setup.

\subsection{LDS evaluation setup}
\label{appendix:b3}
For LDS evaluation, we sample $M{=}64$ random subsets $\mathbb{S}_{m}$ from the training set, each containing $50\%$ of the training data. For each subset $\mathbb{S}_{m}$, three models are trained using different random seeds to reduce evaluation variance. For each input (sample of interest) $\vx$, the linear datamodeling score is computed as the Spearman rank correlation between model outputs and attribution scores, as described in \eqref{eq:lds}. Specifically, we compute the model output $\boldsymbol{\mathcal{F}}=\mathcal{L}_{\text{Simple}}(\vx, \theta)$ in \eqref{eq:simple_objective} over $1{,}000$ timesteps uniformly spaced in the interval $[1, T]$. At each timestep, we approximate the expectation $\mathbb{E}_{\boldsymbol{\epsilon}}$ by sampling three standard Gaussian noise $\boldsymbol{\epsilon} \sim \mathcal{N}(0, \mathbf{I})$. Final LDS performance is reported by averaging scores over both validation and generation samples.

\subsection{Baselines}
\label{appendix:baselines}
We compare NDA against two major categories of attribution methods: (1) \textbf{nonparametric similarity-based methods}, which do not rely on model parameters and instead operate directly on image features and representations, and (2) \textbf{post-hoc retraining-free methods}, which leverage gradient-based representations of a trained model to estimate the attribution score without retraining. Within the first category, we evaluate two similarity-based methods: Raw Pixel and CLIP~\citep{radford2021learning}. Within the second category, we further divide the methods into \emph(gradient-based methods without kernels) and \emph(gradient-based methods with kernels). The first group includes Gradient~\citep{NEURIPS2019_c61f571d}, TracInCP~\citep{NEURIPS2020_e6385d39}, and GAS~\citep{10.1145/3548606.3559335}. The second group includes representative kernel-based methods such as TRAK~\citep{pmlr-v202-park23c} and D-TRAK~\citep{zheng2024intriguing}.
\label{appendix:b4}
\looseness=-1

\textbf{Raw Pixel.} A naive similarity-based method that directly treats the raw image as the feature representation. Attribution scores are computed by measuring similarity (e.g., dot product or cosine similarity) between the query sample and each training sample.

\textbf{CLIP Similarity.} Each target and training sample is embedded using CLIP~\citep{radford2021learning}. Attribution is then computed as the cosine similarity (or dot product of normalized embeddings) between pairs of embeddings. This provides a simple estimate of training data influence, independent of the target diffusion model.

\textbf{Gradient.} A gradient-based influence estimator introduced by \citet{NEURIPS2019_c61f571d}. The attribution score for each training sample is computed as the dot product or cosine similarity between its gradient (w.r.t.\ model parameters) and that of the test sample:
\begin{align*}
\tau(\vx,\vz^{(n)};\mathbb{S})
&=
\mathcal{P}^{\top}\nabla_{\theta}\mathcal{L}_{\text{Simple}}(\vx;\theta^{*})^{\top}.
\mathcal{P}^{\top}\nabla_{\theta}\mathcal{L}_{\text{Simple}}(\vz^{(n)};\theta^{*})\textrm{,}
\\[6pt]
\textrm{or}\quad\tau(\vx,\vz^{(n)};\mathbb{S})
&=
\frac{
\mathcal{P}^{\top}\nabla_{\theta}\mathcal{L}_{\text{Simple}}(\vx;\theta^{*})^{\top}.
\mathcal{P}^{\top}\nabla_{\theta}\mathcal{L}_{\text{Simple}}(\vz^{(n)};\theta^{*})
}{
\left\|\mathcal{P}^{\top}\nabla_{\theta}\mathcal{L}_{\text{Simple}}(\vx;\theta^{*})\right\|
\left\|\mathcal{P}^{\top}\nabla_{\theta}\mathcal{L}_{\text{Simple}}(\vz^{(n)};\theta^{*})\right\|
}\textrm{,}
\end{align*}
where $\mathcal{P}$ is a Gaussian random projection matrix that projects the gradient into low-dimensional subspace.

\textbf{TracInCP.} We adopt the TracInCP estimator introduced by \citet{NEURIPS2020_e6385d39}, formulated as
$\tau(\vx, \vz^{(n)}; \mathbb{S})=\frac{1}{C} \sum_{c=1}^C\left(\mathcal{P}_c^{\top} \nabla_\theta \mathcal{L}_{\text {Simple}}\left(\vx; \theta^c\right)\right)^{\top} \cdot\left(\mathcal{P}_c^{\top} \nabla_\theta \mathcal{L}_{\text{Simple}}\left(\vz^{(n)}; \theta^c\right)\right)$,
where $C$ denotes the number of uniformly sampled checkpoints along the training trajectory, and $\theta^{c}$ denotes parameters at the $c$-th checkpoint. We use four checkpoints for each experiment. For example, we select checkpoints at epochs $\{50,100,150,200\}$ on CIFAR-2.

\textbf{GAS.} A ``renormalized'' version of TracInCP that replaces dot products with cosine similarity~\citep{10.1145/3548606.3559335}.

\textbf{TRAK.} TRAK~\citep{pmlr-v202-park23c} can be extended to diffusion models following \citet{zheng2024intriguing}.
The retraining-free TRAK can be implemented as:
$$
\begin{aligned}
& \Phi_{\text{TRAK}}=\left[\phi\left(\vx^1\right), \cdots, \phi\left(\vx^N\right)\right]^{\top} \textrm{, where} \ \phi(\vx)=\mathcal{P}^{\top} \nabla_\theta \mathcal{L}_{\text{Simple}}(\vx; \theta^{*}); \\
& \tau(\vx, \vz^{(n)}; \mathbb{S})=\mathcal{P}^{\top} \nabla_\theta \mathcal{L}_{\text{Simple}}(\vx; \theta^{*})^{\top} \cdot\left(\Phi_{\text{TRAK}}{ }^{\top} \Phi_{\text{TRAK}}+\lambda I\right)^{-1} \cdot \mathcal{P}^{\top} \nabla_\theta \mathcal{L}_{\text{Simple}}(\vz^{(n)}; \theta^{*})\textrm{,}
\end{aligned}
$$
where the term $\lambda I$ is included for numerical stability and regularization effect.

\textbf{D-TRAK}~\citep{zheng2024intriguing}. Similar to TRAK but uses the squared-error objective:
$$
\begin{aligned}
& \Phi_{\text{D-TRAK}}=\left[\phi\left(\vx^1\right), \cdots, \phi\left(\vx^N\right)\right]^{\top} \textrm{, where} \ \phi(\vx)=\mathcal{P}^{\top} \nabla_\theta \mathcal{L}_{\text {Square}}(\vx; \theta^{*}); \\
& \tau(\vx, \vz^{(n)}; \mathbb{S})=\left(\mathcal{P}^{\top} \nabla_\theta \mathcal{L}_{\text {Square}}(\vx ; \theta^{*})\right)^{\top} \cdot\left(\Phi_{\text{D-TRAK}}{ }^{\top} \Phi_{\text{D-TRAK}}+\lambda I\right)^{-1} \cdot \mathcal{P}^{\top} \nabla_\theta \mathcal{L}_{\text{Square}}(\vz^{(n)}; \theta^{*})\textrm{,}
\end{aligned}
$$
with $\mathcal{L}_{\text{Square}} = \|\boldsymbol{\epsilon}_{\theta}(\vx_{t}, t)\|^{2}$.

\section{Additional ablation studies}
\label{appendix:abla}

We provide additional ablation studies on patch size selection and multiscale influence.

\begin{figure}[t]
    \centering
    \subfloat{\includegraphics[width=0.33\textwidth]{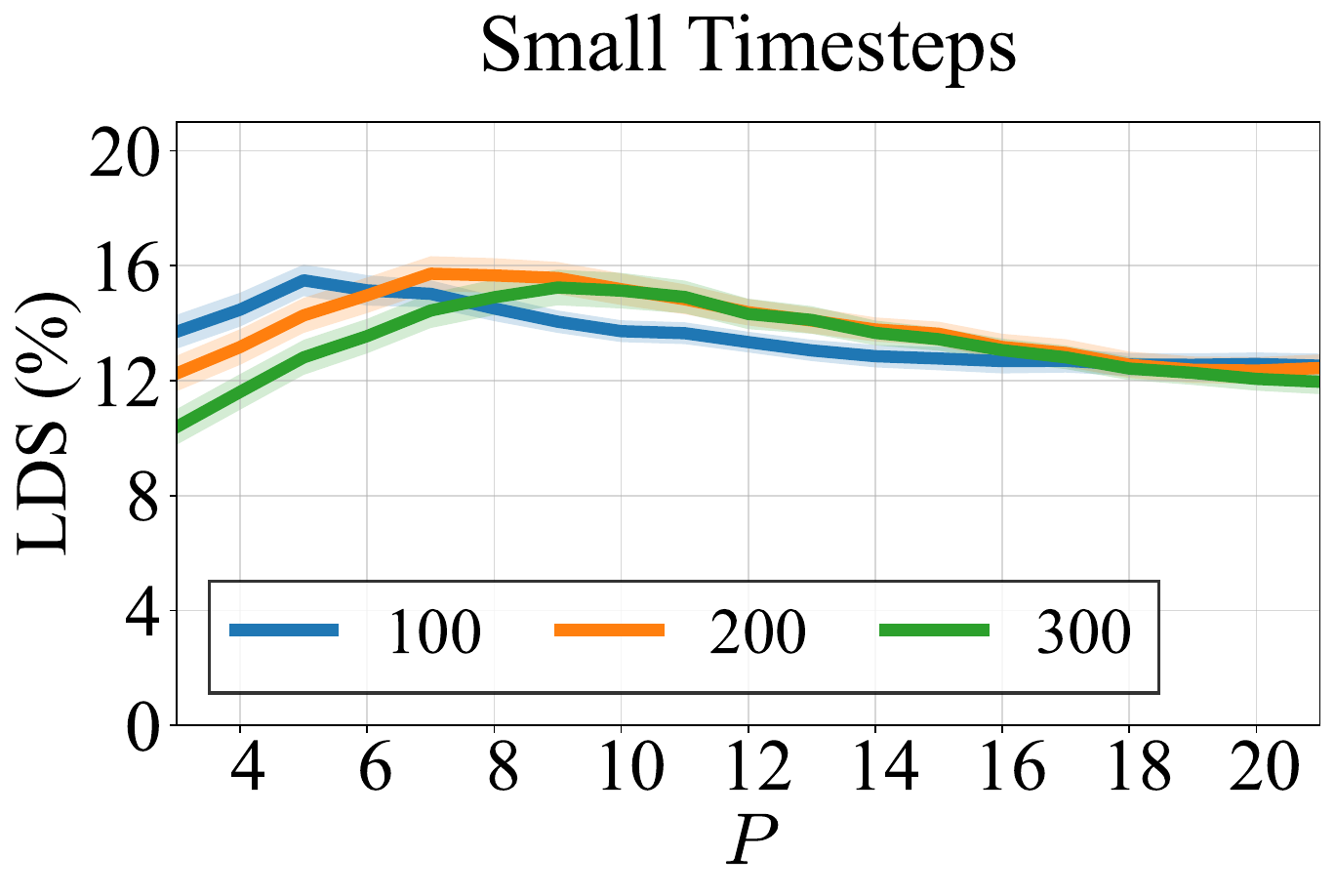}}
    \subfloat{\includegraphics[width=0.33\textwidth]{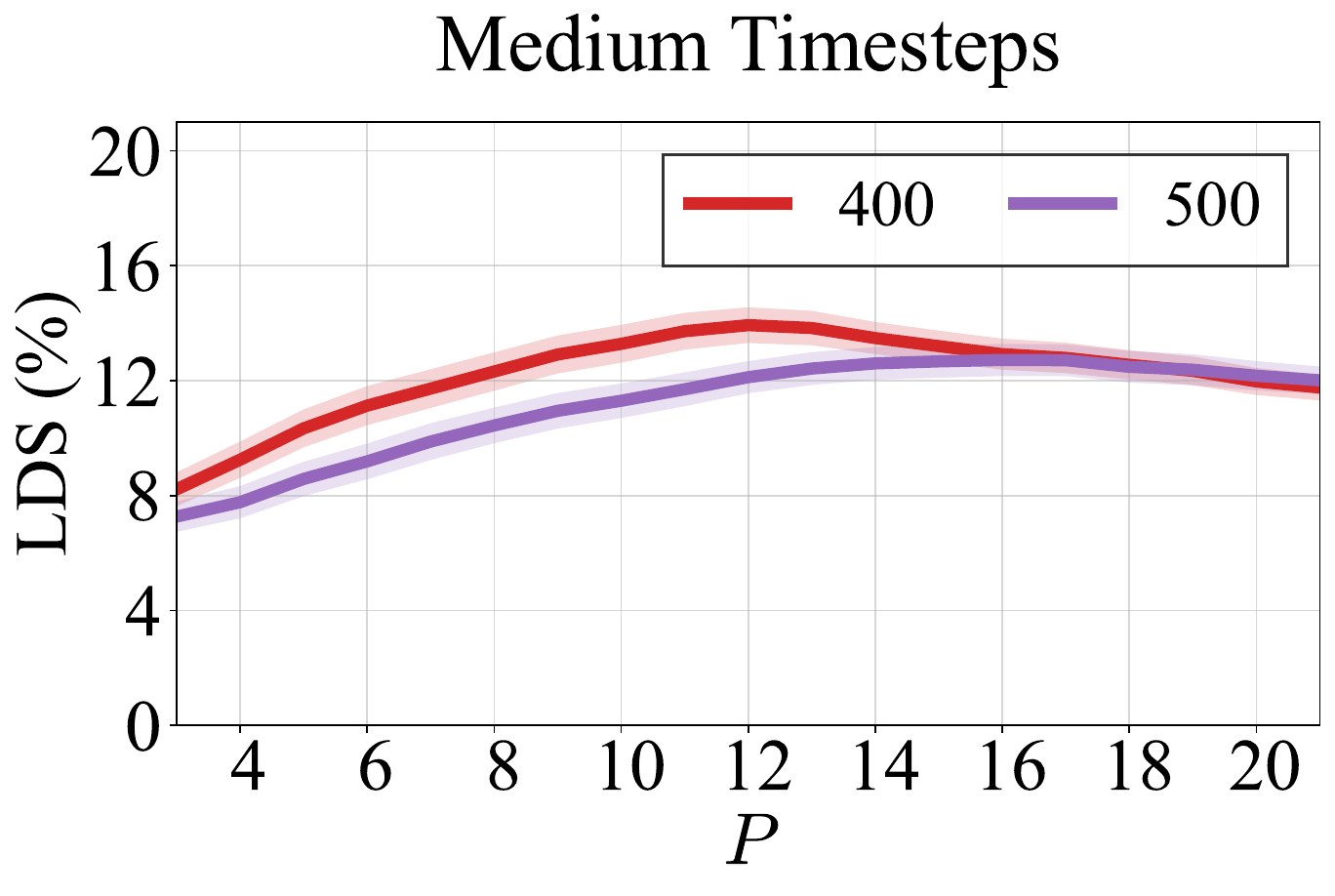}}
    \subfloat{\includegraphics[width=0.33\textwidth]{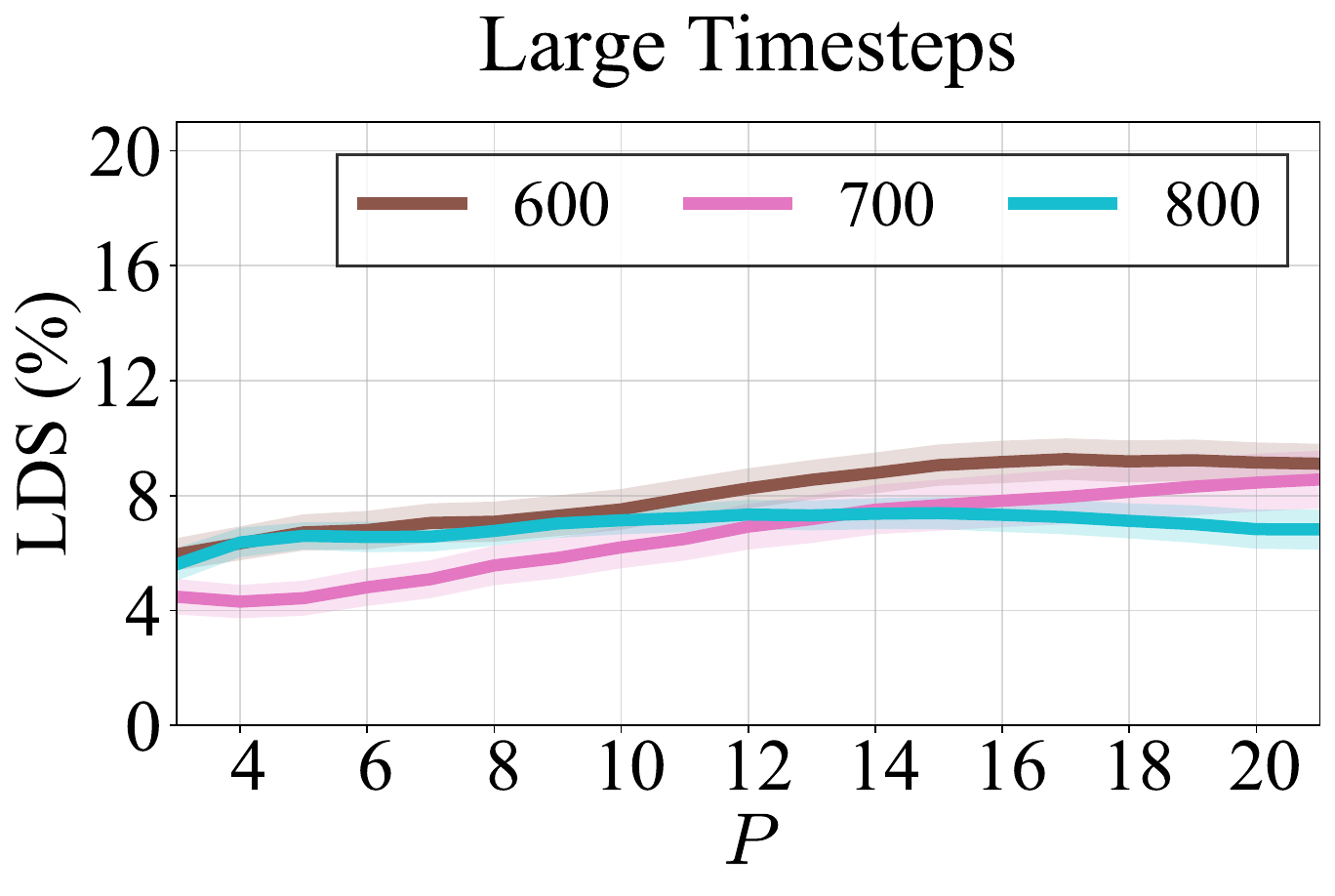}}
\caption{
LDS (\%) on the CelebA validation set across different patch sizes and timesteps at the original resolution. \textbf{Left}: small timesteps ($t{\leq}300$), where moderate patch sizes achieve the highest scores. \textbf{Middle}: medium timesteps ($t{=}400,500$), where larger patches outperform smaller ones due to higher noise. \textbf{Right}: large timesteps, where noise dominates and informative signal is minimal.\looseness=-1
}
\label{fig:celeba_ps-vs-timestep}
\end{figure}

\begin{figure}[t]
    \centering
    \subfloat{\includegraphics[width=0.33\textwidth]{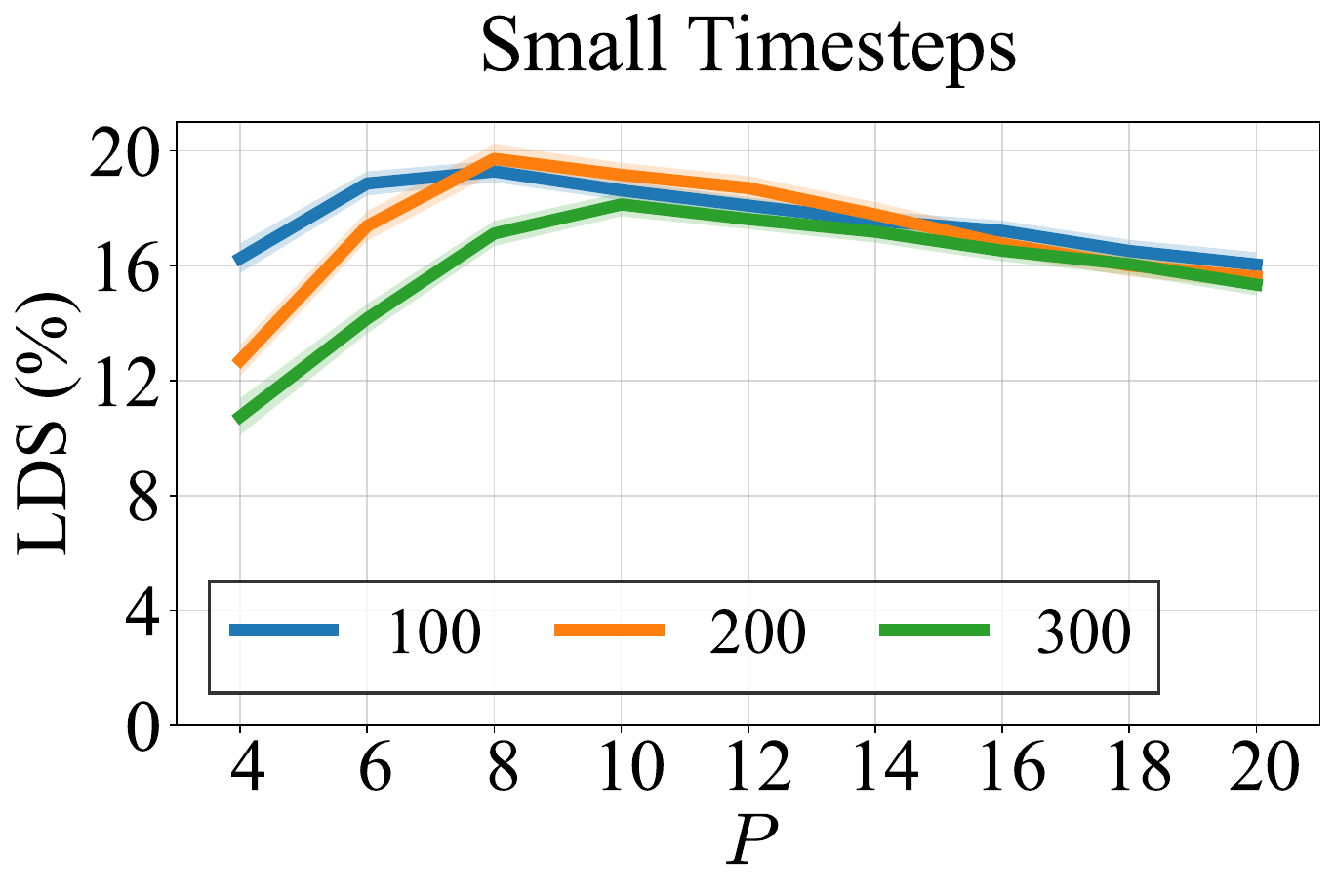}}
    \subfloat{\includegraphics[width=0.33\textwidth]{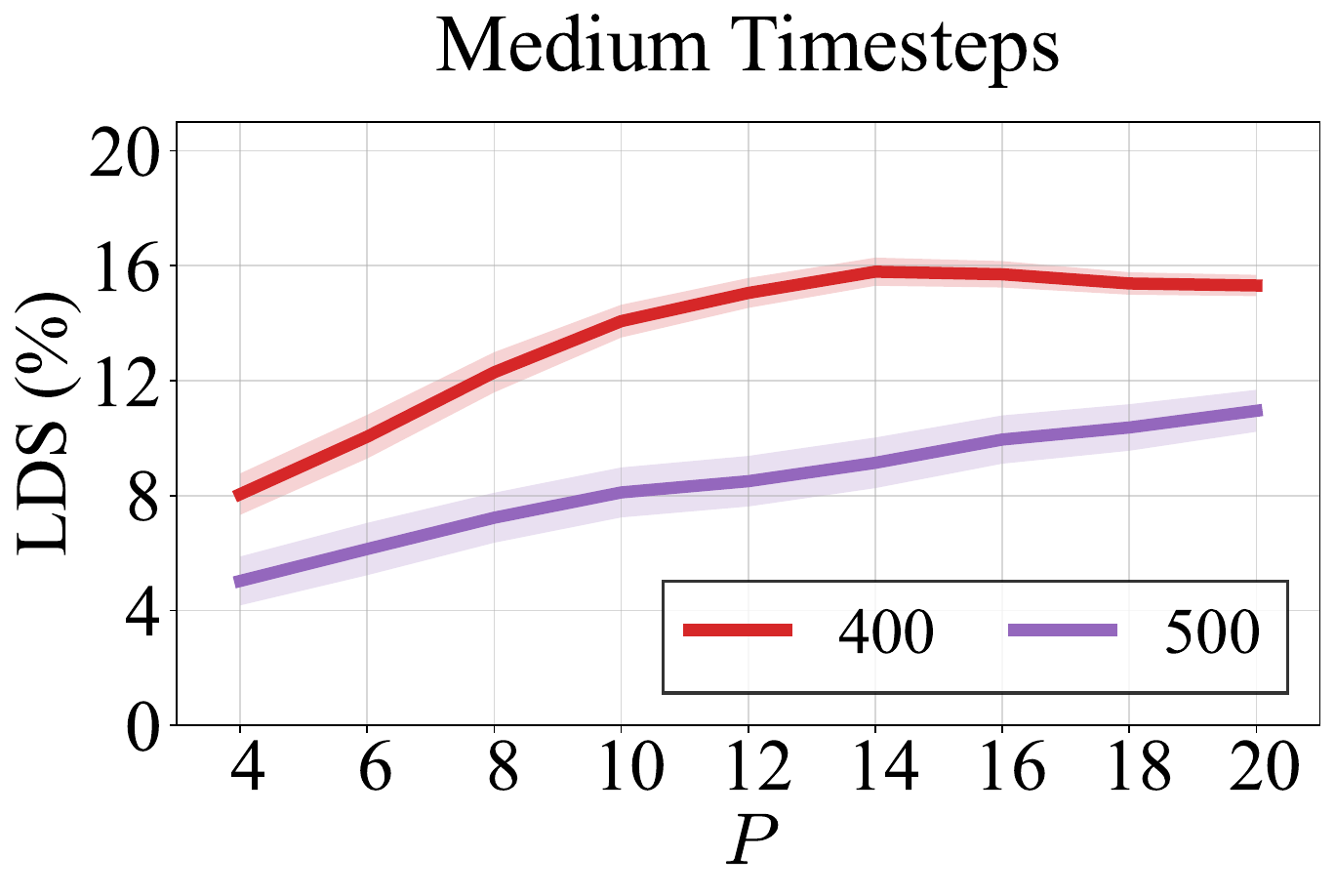}}
    \subfloat{\includegraphics[width=0.33\textwidth]{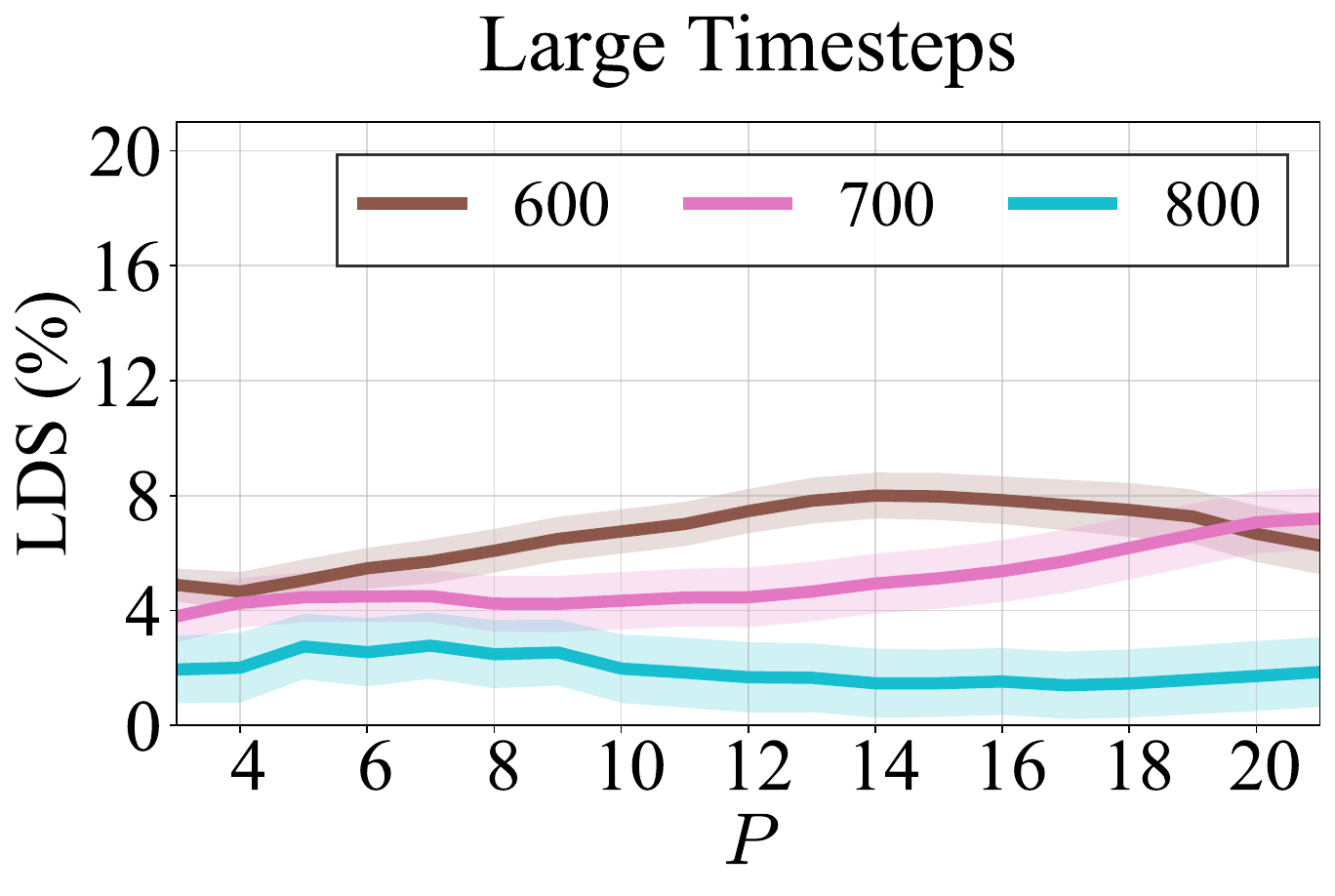}}
    \caption{LDS (\%) on the CIFAR-2 validation set across different patch sizes and timesteps at the low resolution. \textbf{Left}: small timesteps ($t{\leq}300$), where moderate patch sizes achieve the highest scores. \textbf{Middle}: medium timesteps ($t{=}400,500$), where larger patches outperform smaller ones due to higher noise. \textbf{Right}: large timesteps, where noise dominates and informative signal is minimal.\looseness=-1}
\label{fig:cifar2_ps-vs-timestep_downscale}
\end{figure}

\begin{figure}[t]
    \centering
    \subfloat{\includegraphics[width=0.33\textwidth]{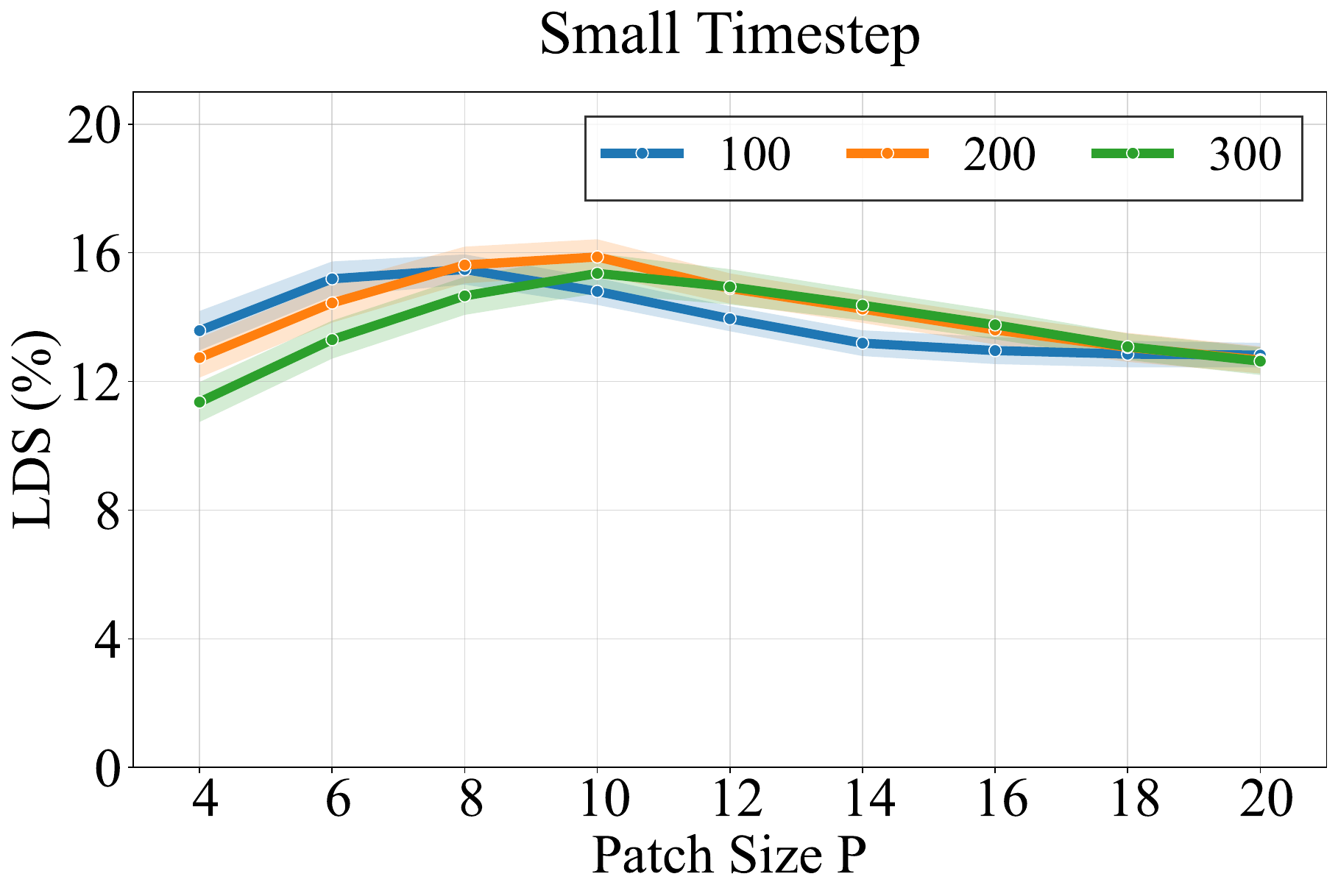}}
    \subfloat{\includegraphics[width=0.33\textwidth]{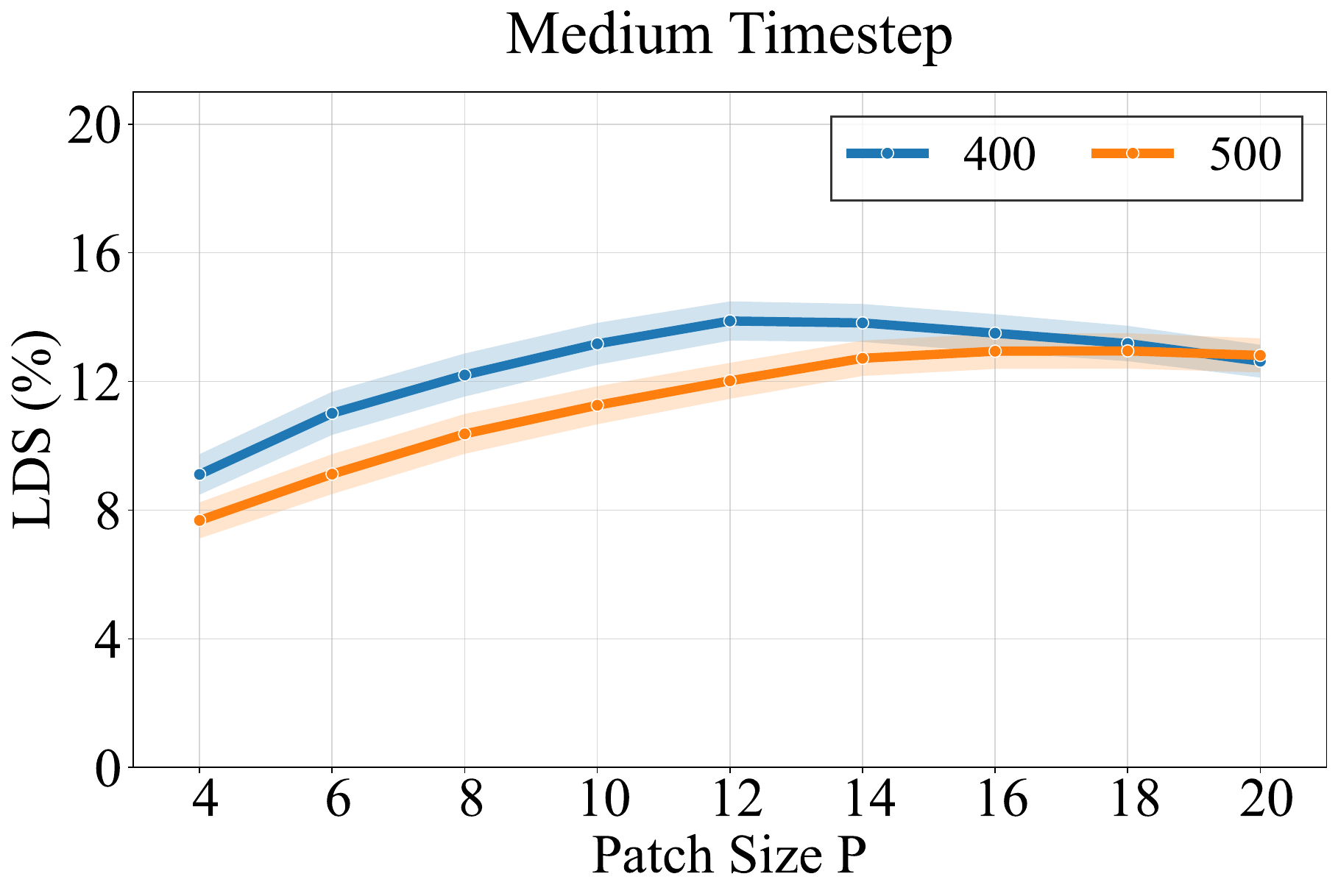}}
    \subfloat{\includegraphics[width=0.33\textwidth]{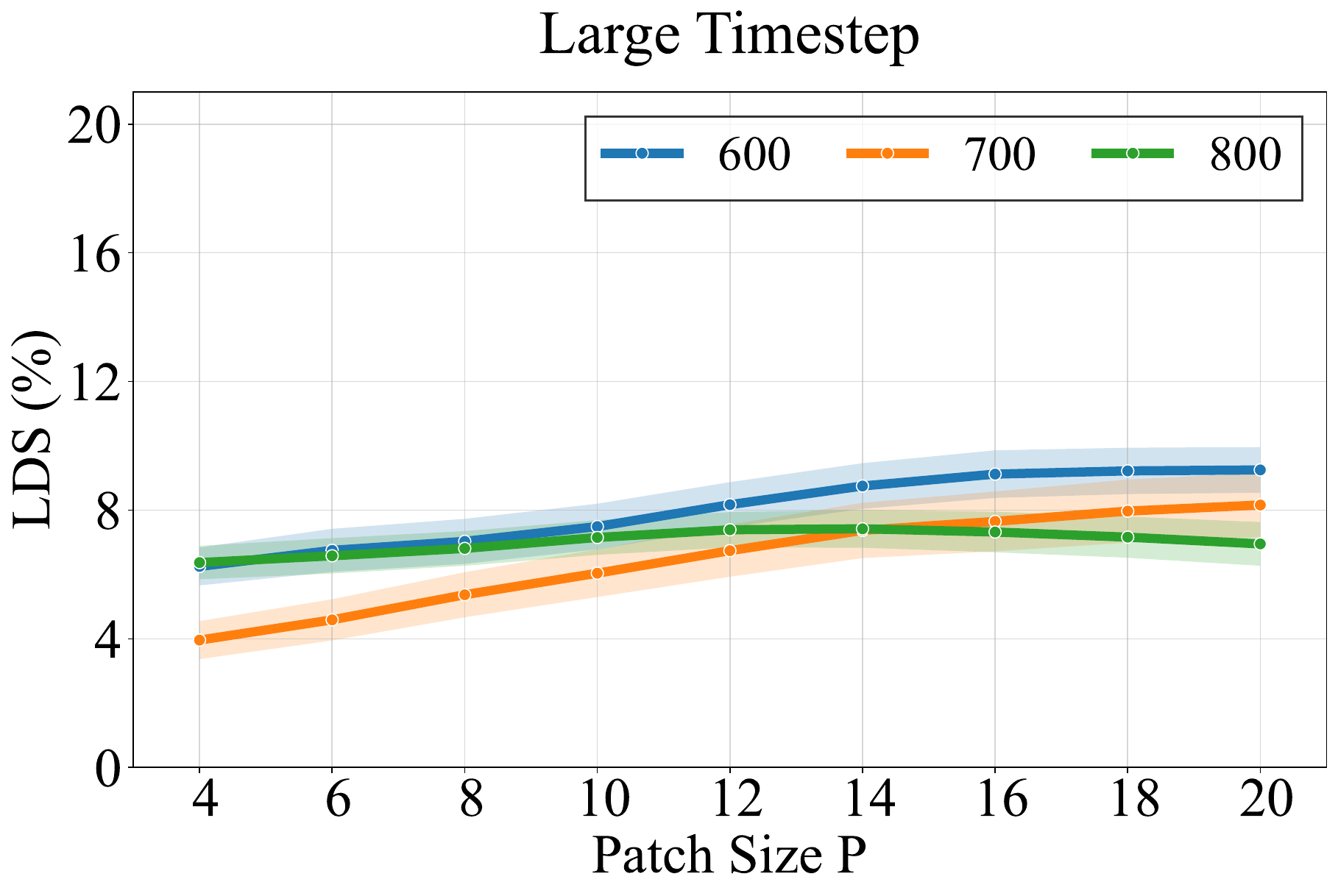}}
    \caption{
LDS (\%) on the CelebA validation set across different patch sizes and timesteps at the low resolution. \textbf{Left}: small timesteps ($t{\leq}300$), where moderate patch sizes achieve the highest scores. \textbf{Middle}: medium timesteps ($t{=}400,500$), where larger patches outperform smaller ones due to higher noise. \textbf{Right}: large timesteps, where noise dominates and informative signal is minimal.\looseness=-1
}
\label{fig:celeba_ps-vs-timestep-down}
\end{figure}

\textbf{Patch size selection.} For completeness, we also study the effect of patch size by evaluating LDS across \(P\in[3,21]\) at different timesteps on CelebA. To isolate timestep effects, we fix $\mathcal{T}$ to a single $t$ and vary $P$. Figure~\ref{fig:celeba_ps-vs-timestep} shows LDS as a function of $P$ for each $t$. The trend mirrors CIFAR-2: the optimal patch size generally grows with $t$. At eraly timesteps ($t{\le} 300$), small to moderate patches (e.g., $P{=} 5,7,9$) yield the highest scores, indicating local patterns dominate when noise is low. At mid-range timesteps ($400{\le} t{\le} 500$), larger patches perform better (peaks shift to $P{=} 11,13$), likely due to the need for more contextual information under higher noise. In the high-noise regime ($t{\ge} 600$), LDS drops and curves flatten across all patch sizes, suggesting informative signal is limited.

\textbf{Multiscale influence.} We further compare the optimal patch size between the original and the low-resolution settings on CIFAR-2. As shown in Figure~\ref{fig:cifar2_ps-vs-timestep} and Figure~\ref{fig:cifar2_ps-vs-timestep_downscale}, at the early timestep \(t{=}100\), the optima differ notably: the original resolution peaks at $P{=}5$, whereas the low-resolution curve peaks at $P{=}8$, indicating that coarser inputs benefit from slightly larger spatial context. By contrast, at $t{=}200$ and $t{=}300$, the two resolutions behave very similarly: on the original scale it peaks at $P{=}7$ and $P{=}9$ while low-resolution peaks at $P{=}8$ and $P{=}10$, with near-overlapping curves around their maxima. At mid timesteps ($t{=}400, 500$), both resolution favors large patch size with $P{=}21$. These results show that downscaling mainly shifts optimal to large patch size $P$ at small timestep, while mid/large timesteps show qualitatively similar trends across resolutions.

\section{Visualization}
\label{appendix:vis}
We provide additional visualizations, including counterfactual examples and proponent-opponent analyses.
\subsection{Counterfactual visulization}
\label{appendix:vis_counter}
\begin{figure}[ht]
    \centering
    \scriptsize
    \makebox[\textwidth]{ 
    \begin{tabular}{c@{\hspace{2em}}c}
        \begin{tabular}{@{}c@{}c@{}c@{}c@{}c@{}}
             & \textbf{Random} & \textbf{CLIP} & \textbf{Ours} & \textbf{D-TRAK} \\
            \includegraphics[width=0.1\textwidth]{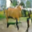} &
            \includegraphics[width=0.1\textwidth]{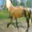} &
            \includegraphics[width=0.1\textwidth]{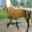} &
            \includegraphics[width=0.1\textwidth]{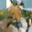} &
            \includegraphics[width=0.1\textwidth]{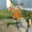} \\
            \vspace{0.1em} \\
            \includegraphics[width=0.1\textwidth]{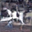} &
            \includegraphics[width=0.1\textwidth]{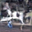} &
            \includegraphics[width=0.1\textwidth]{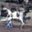} &
            \includegraphics[width=0.1\textwidth]{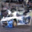} &
            \includegraphics[width=0.1\textwidth]{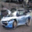} \\
            \vspace{0.1em} \\
            \includegraphics[width=0.1\textwidth]{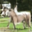} &
            \includegraphics[width=0.1\textwidth]{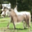} &
            \includegraphics[width=0.1\textwidth]{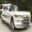} &
            \includegraphics[width=0.1\textwidth]{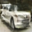} &
            \includegraphics[width=0.1\textwidth]{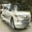}
        \end{tabular}
        &
        \begin{tabular}{@{}c@{}c@{}c@{}c@{}c@{}}
             & \textbf{Random} & \textbf{CLIP} & \textbf{Ours} & \textbf{D-TRAK} \\
            \includegraphics[width=0.1\textwidth]{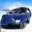} &
            \includegraphics[width=0.1\textwidth]{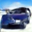} &
            \includegraphics[width=0.1\textwidth]{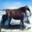} &
            \includegraphics[width=0.1\textwidth]{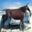} &
            \includegraphics[width=0.1\textwidth]{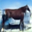} \\
            \vspace{0.1em} \\
            \includegraphics[width=0.1\textwidth]{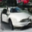} &
            \includegraphics[width=0.1\textwidth]{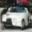} &
            \includegraphics[width=0.1\textwidth]{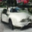} &
            \includegraphics[width=0.1\textwidth]{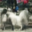} &
            \includegraphics[width=0.1\textwidth]{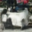} \\
            \vspace{0.1em} \\
            \includegraphics[width=0.1\textwidth]{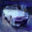} &
            \includegraphics[width=0.1\textwidth]{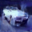} &
            \includegraphics[width=0.1\textwidth]{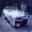} &
            \includegraphics[width=0.1\textwidth]{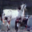} &
            \includegraphics[width=0.1\textwidth]{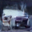}
        \end{tabular}
    \end{tabular}
    }
    \caption{Counterfactual visualization on CIFAR-2 dataset. We compare samples generated by retrained models with different attribution methods using the same seed.}
    \label{fig:counter_vis2}
\end{figure}

\begin{figure}[ht]
    \centering
    \scriptsize
    \makebox[\textwidth]{ 
    \begin{tabular}{c@{\hspace{2em}}c}
        \begin{tabular}{@{}c@{}c@{}c@{}c@{}c@{}}
             & \textbf{Random} & \textbf{CLIP} & \textbf{Ours} & \textbf{D-TRAK} \\
             \includegraphics[width=0.1\textwidth]{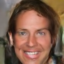} &
            \includegraphics[width=0.1\textwidth]{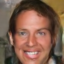} &
            \includegraphics[width=0.1\textwidth]{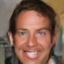} &
            \includegraphics[width=0.1\textwidth]{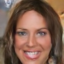} &
            \includegraphics[width=0.1\textwidth]{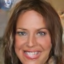} \\    
            \vspace{0.1em} \\
            \includegraphics[width=0.1\textwidth]{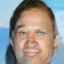} &
            \includegraphics[width=0.1\textwidth]{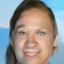} &
            \includegraphics[width=0.1\textwidth]{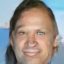} &
            \includegraphics[width=0.1\textwidth]{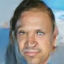} &
            \includegraphics[width=0.1\textwidth]{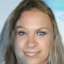} \\
            \vspace{0.1em} \\
            \includegraphics[width=0.1\textwidth]{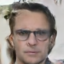} &
            \includegraphics[width=0.1\textwidth]{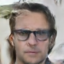} &
            \includegraphics[width=0.1\textwidth]{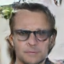} &
            \includegraphics[width=0.1\textwidth]{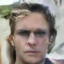} &
            \includegraphics[width=0.1\textwidth]{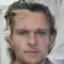}
        \end{tabular}
        &
        \begin{tabular}{@{}c@{}c@{}c@{}c@{}c@{}}
             & \textbf{Random} & \textbf{CLIP} & \textbf{Ours} & \textbf{D-TRAK} \\
            \includegraphics[width=0.1\textwidth]{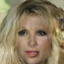} &
            \includegraphics[width=0.1\textwidth]{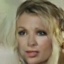} &
            \includegraphics[width=0.1\textwidth]{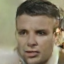} &
            \includegraphics[width=0.1\textwidth]{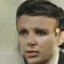} &
            \includegraphics[width=0.1\textwidth]{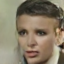} \\
            \vspace{0.1em} \\
            \includegraphics[width=0.1\textwidth]{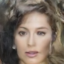} &
            \includegraphics[width=0.1\textwidth]{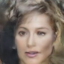} &
            \includegraphics[width=0.1\textwidth]{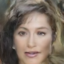} &
            \includegraphics[width=0.1\textwidth]{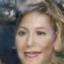} &
            \includegraphics[width=0.1\textwidth]{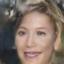} \\
            \vspace{0.1em} \\
            \includegraphics[width=0.1\textwidth]{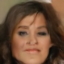} &
            \includegraphics[width=0.1\textwidth]{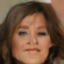} &
            \includegraphics[width=0.1\textwidth]{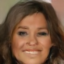} &
            \includegraphics[width=0.1\textwidth]{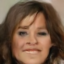} &
            \includegraphics[width=0.1\textwidth]{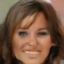}
        \end{tabular}
    \end{tabular}
    }
    \caption{Counterfactual visualization on CelebA dataset. We compare samples generated by retrained models with different attribution methods using the same seed.}
    \label{fig:counter_vis3}
\end{figure}

We include more counterfactual visualizations in Figures~\ref{fig:counter_vis2} and \ref{fig:counter_vis3}. As shown, NDA identifies training examples whose removal leads to marked changes in the output of a model retrained with the same seed. In Figure~\ref{fig:counter_vis2} (right column, third row), models retrained after removing images selected by Random and CLIP still synthesize an ``automobile''. In contrast, removing images selected by NDA or D-TRAK yields an image that resembles a mixture of ``automobile'' and ``horse''. Notably, the image after NDA’s removal shows clearer horse morphology (e.g., head and leg outlines) than that after D-TRAK. These results suggest that NDA is an effective nonparametric approach for identifying training images with strong influence on a given target.

\subsection{Proponents and opponents visulization}
\begin{figure}[htbp]
    \centering

    \includegraphics[width=0.95\textwidth]{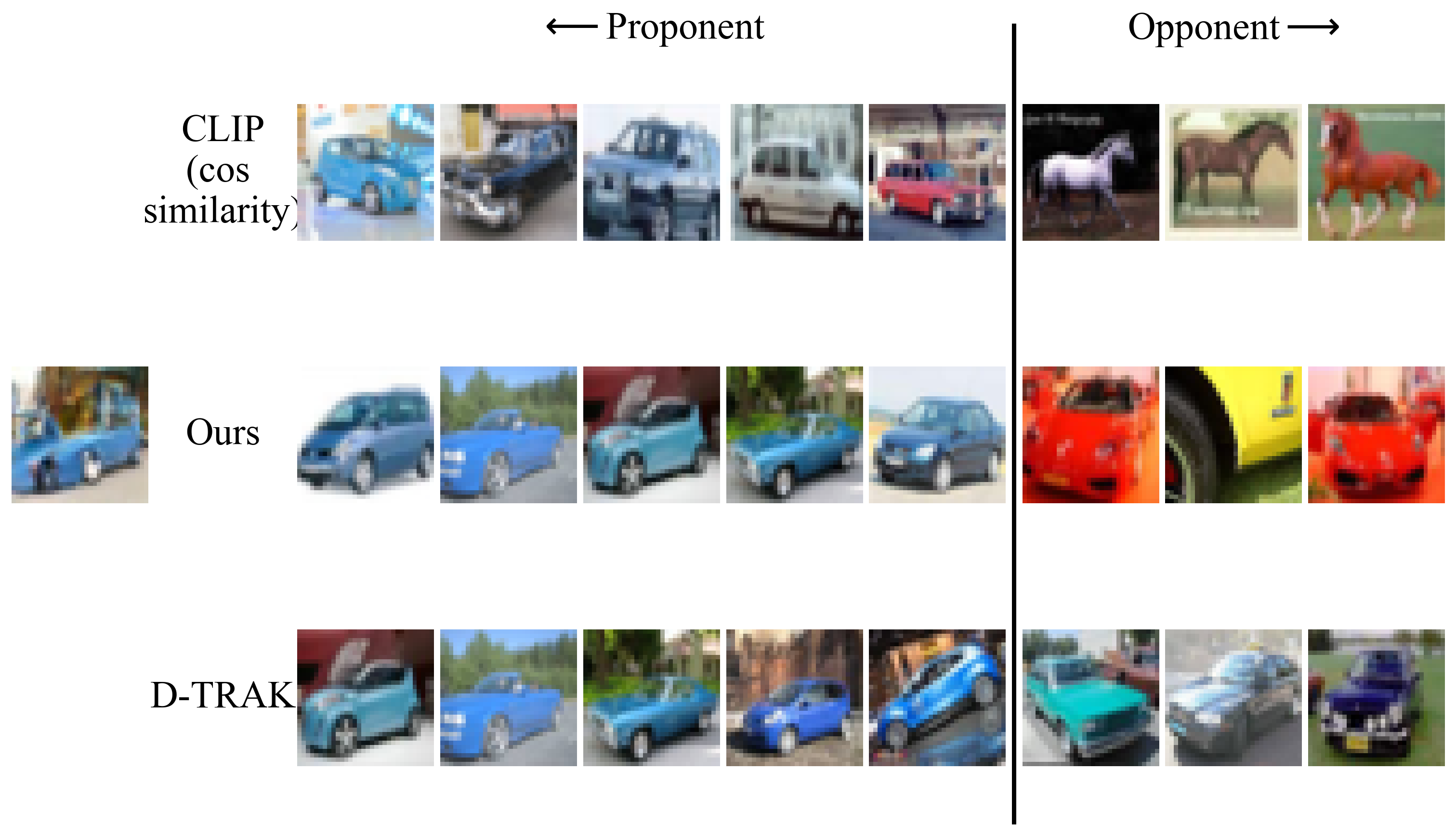}
    
    \vspace{0.5ex}
    \makebox[0.95\textwidth][c]{(a) Generation}
    
    \vspace{2ex}

    \includegraphics[width=0.95\textwidth]{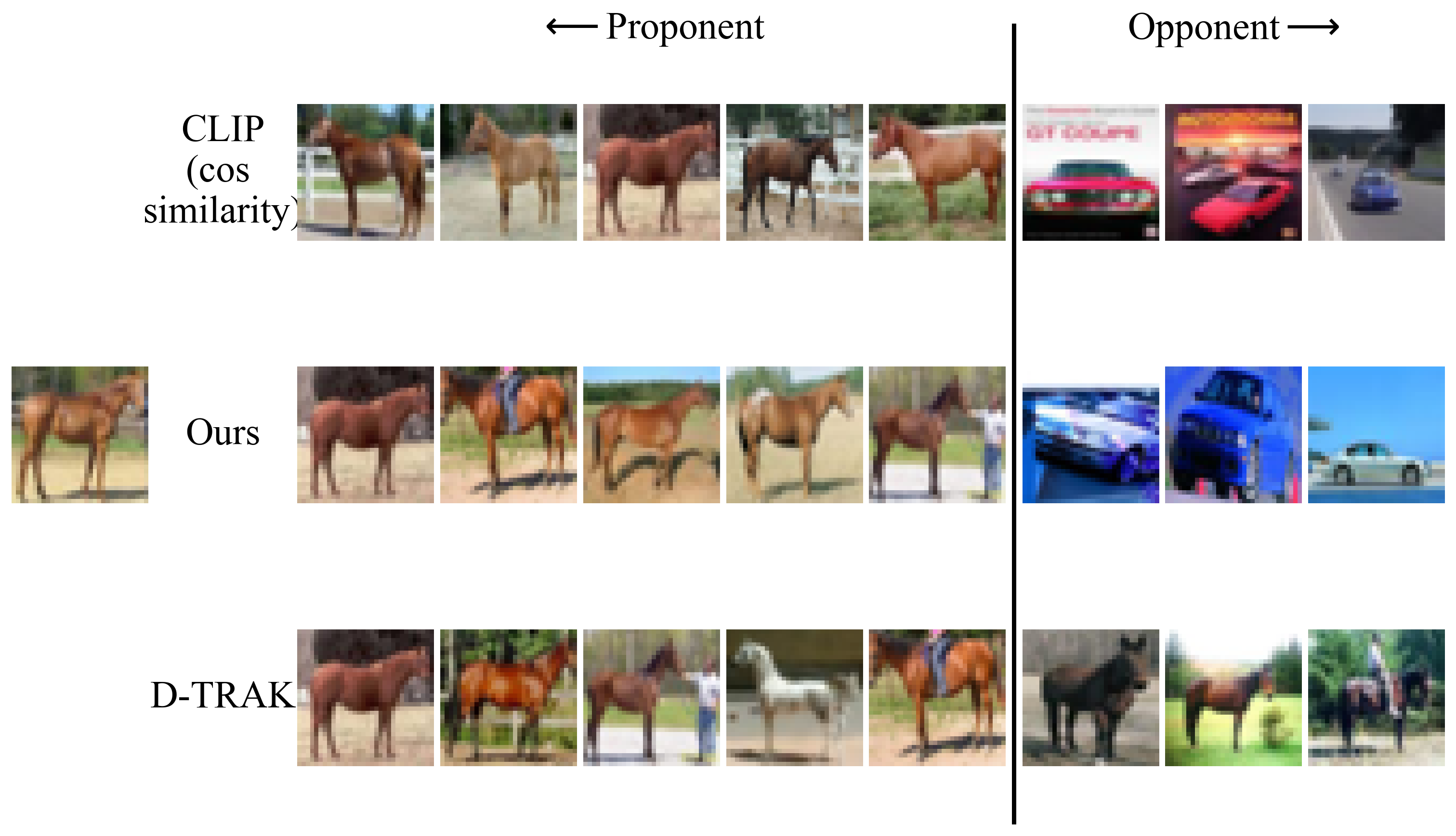}
    
    \vspace{0.5ex}
    \makebox[0.95\textwidth][c]{(b) Validation}

    \caption{
        Proponents and opponents visualization on CIFAR-2 using CLIP, NDA and D-TRAK (using averaged timesteps).
        For each target sample, the 5 most positively influential training samples (Left) and the 3 most negatively influential samples (Right) are shown.
    }
    \label{fig:vis_attribution_cifar2}
\end{figure}

\begin{figure}[htbp]
    \centering

    \includegraphics[width=0.95\textwidth]{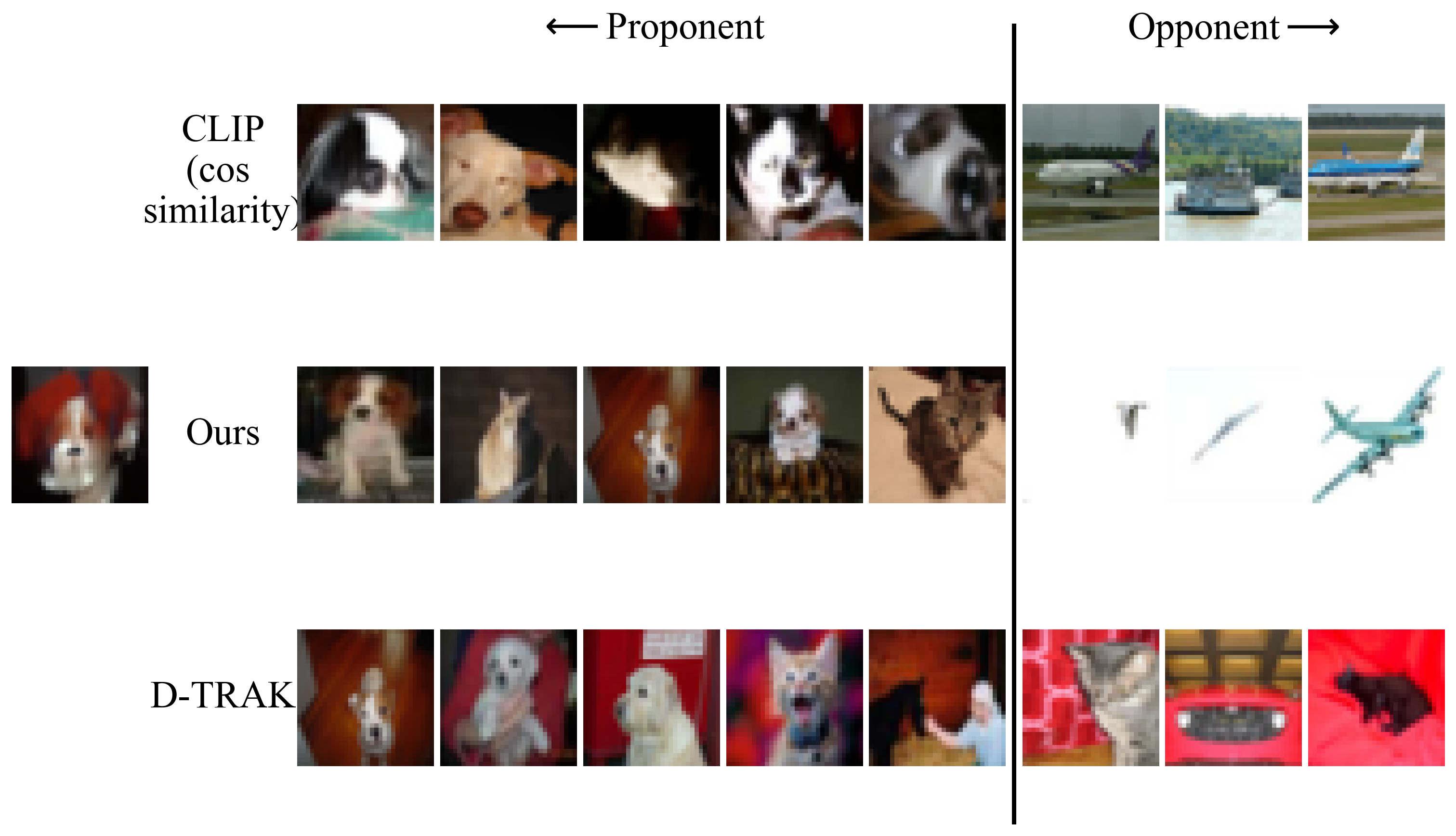}
    
    \vspace{0.5ex}
    \makebox[0.95\textwidth][c]{(a) Generation}
    
    \vspace{2ex}

    \includegraphics[width=0.95\textwidth]{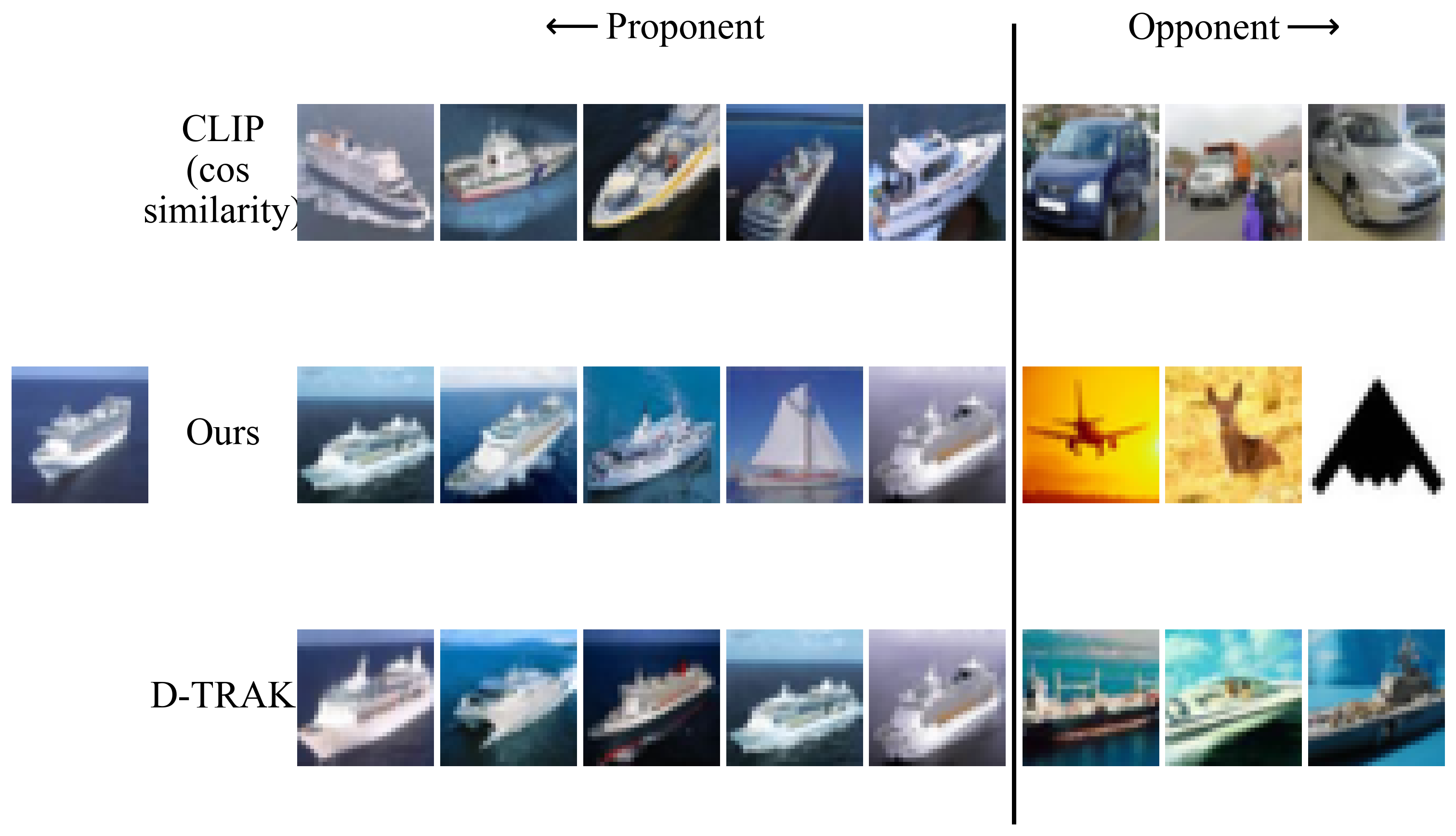}
    
    \vspace{0.5ex}
    \makebox[0.95\textwidth][c]{(b) Validation}

    \caption{
        Proponents and opponents visualization on CIFAR-10 using CLIP, NDA and D-TRAK (using averaged timesteps).
        For each target sample, the 5 most positively influential training samples (Left) and the 3 most negatively influential samples (Right) are shown.
    }
    \label{fig:vis_attribution_cifar10}
\end{figure}

\begin{figure}[htbp]
    \centering

    \includegraphics[width=0.95\textwidth]{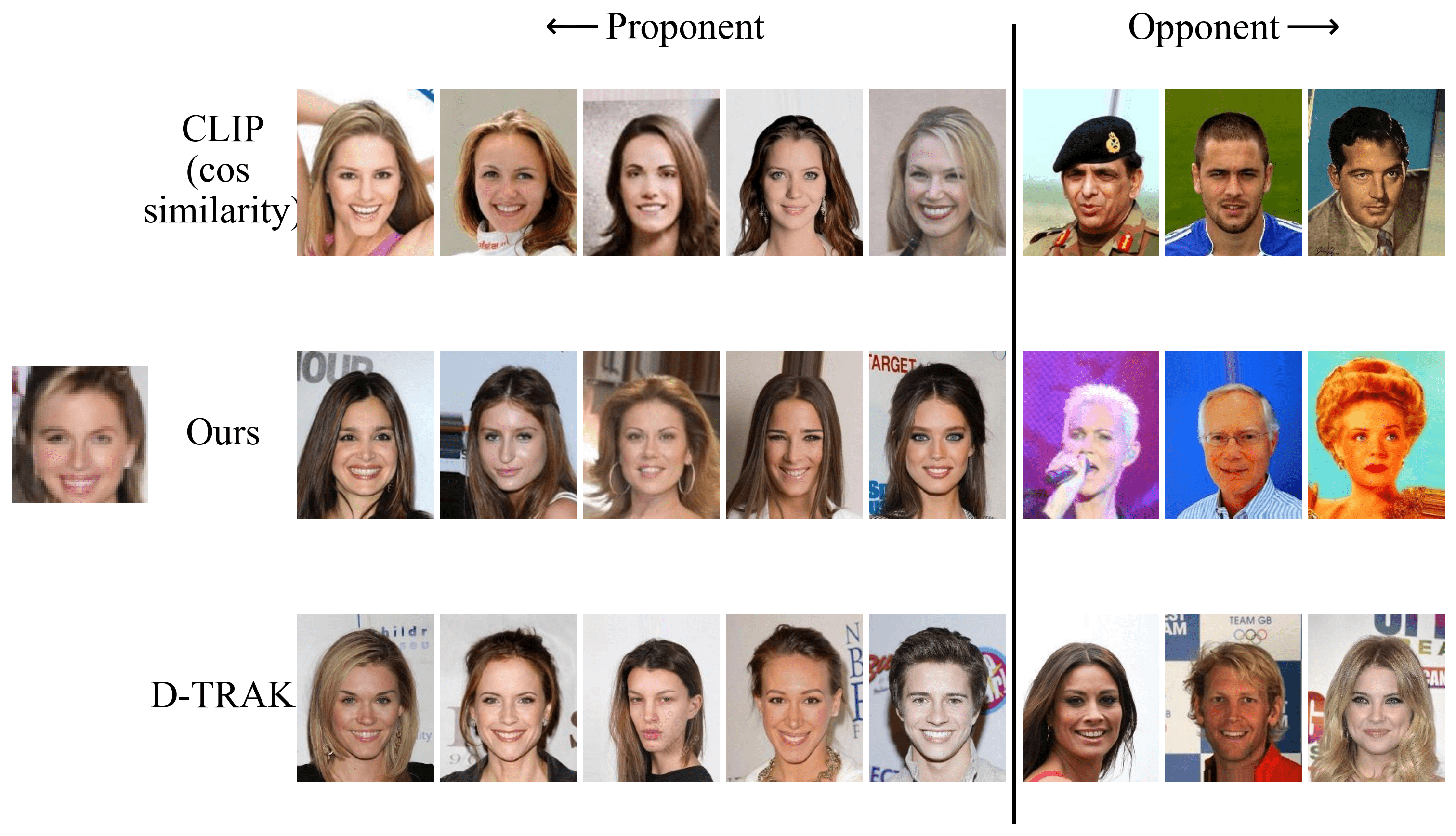}
    
    \vspace{0.5ex}
    \makebox[0.95\textwidth][c]{(a) Generation}
    
    \vspace{2ex}

    \includegraphics[width=0.95\textwidth]{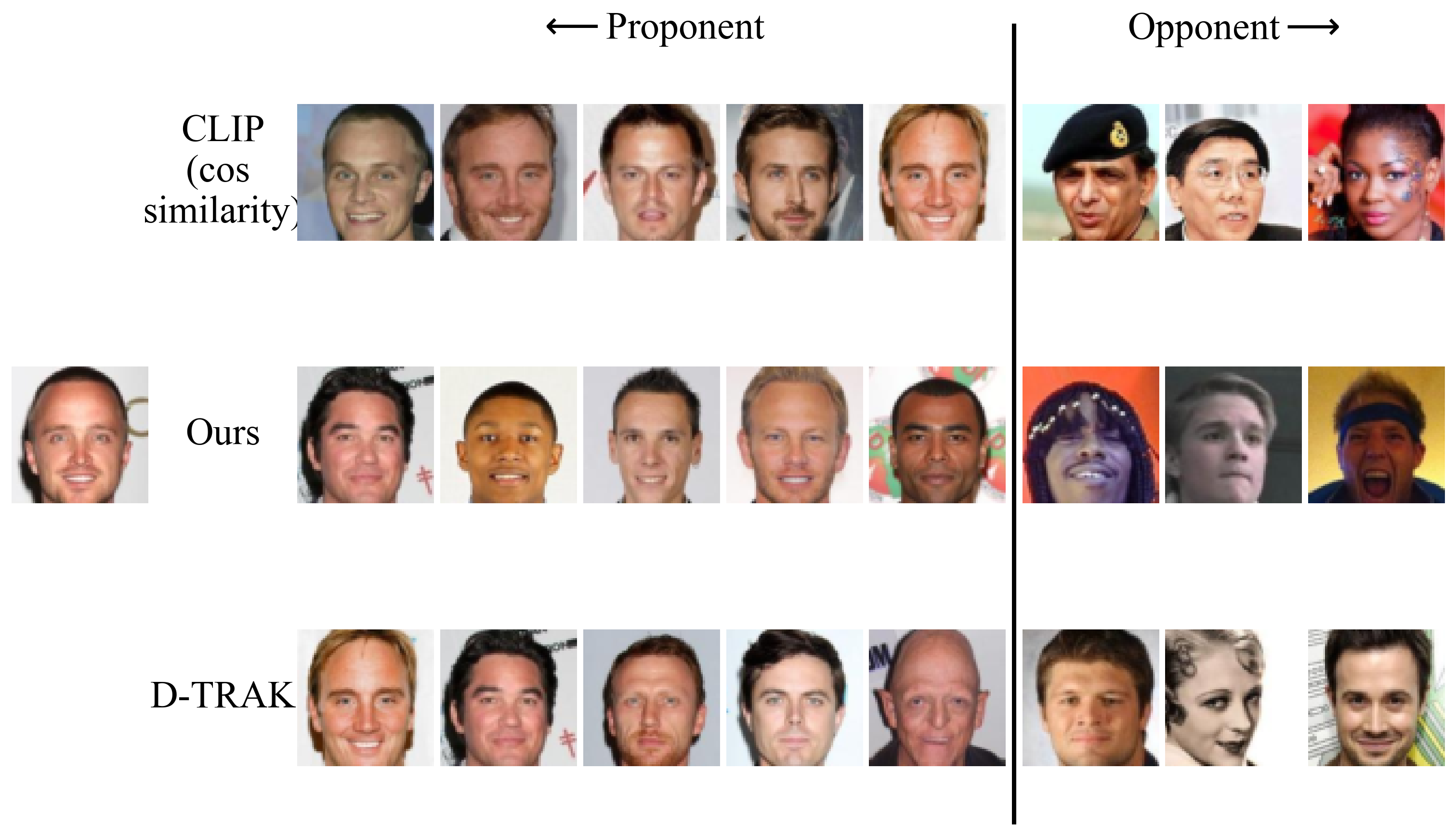}
    
    \vspace{0.5ex}
    \makebox[0.95\textwidth][c]{(b) Validation}

    \caption{
        Proponents and opponents visualization on CelebA using CLIP, NDA and D-TRAK (using averaged timesteps).
        For each target sample, the 5 most positively influential training samples (Left) and the 3 most negatively influential samples (Right) are shown.
    }
    \label{fig:vis_attribution_celebA}
\end{figure}

Following \citet{NEURIPS2020_e6385d39}, we refer to training examples with positive influence scores as \emph{proponents} and those with negative scores as \emph{opponents}. For each target, we retrieve and visualize the top-$5$ proponents and the top-$3$ opponents. Qualitative results on CIFAR-2, CIFAR-10, and CelebA are shown in Figures~\ref{fig:vis_attribution_cifar2}, \ref{fig:vis_attribution_cifar10}, and \ref{fig:vis_attribution_celebA}. Across datasets, NDA consistently retrieves proponents that are visually more similar to the target than those selected by CLIP, and its opponents are correspondingly more dissimilar.

\end{document}